\documentclass[a4paper,11pt]{article}
\usepackage{amsmath,amssymb,amsthm,bbm,bm}  
\usepackage{mathrsfs}
\usepackage{authblk}  
\usepackage[british]{babel}
\usepackage{balance}
\usepackage[utf8]{inputenc}

\usepackage{booktabs} 
\usepackage{caption}
\usepackage{color}
\usepackage{comment}
\usepackage{dsfont}
\usepackage{enumitem}
\usepackage[T1]{fontenc}
\usepackage[top=2.5cm, bottom=2.5cm, left=2.5cm, right=2.5cm]{geometry}
\usepackage{graphicx}
\usepackage{graphbox}       
\PassOptionsToPackage{hyphens}{url}  
\usepackage[colorlinks = true,
            linkcolor = OliveGreen,
            urlcolor  = blue,
            citecolor = MidnightBlue,
            anchorcolor = blue]{hyperref}
\usepackage{cleveref}
\usepackage{csquotes}  
\usepackage[mono=false]{libertine}  
\usepackage{mathtools}
\usepackage{microtype}      
\usepackage{multirow}    
\usepackage{nicefrac}
\usepackage{physics}
\usepackage{subfigure}
\usepackage{wrapfig}
\usepackage[usenames,dvipsnames]{xcolor}

\newcommand{\identity}{\text{Id}}
\newcommand{\stdGauss}{{\mathcal{N}\left(0,\identity\right)}}
\newcommand{\T}{{\mathrm{T}}}

\title{Generalization Dynamics of Linear Diffusion Models}

\author{Claudia Merger}
\author{Sebastian Goldt\thanks{\{cmerger, sgoldt\}@sissa.it}}
\affil{International School of Advanced Studies (SISSA), Trieste, Italy}

\date{\today}
\begin{document}

\maketitle

\begin{abstract}
  \noindent 
Diffusion models are powerful generative models that produce high-quality samples from complex data. While their infinite-data behavior is well understood, their generalization with finite data remains less clear. Classical learning theory predicts that generalization occurs at a sample complexity that is exponential in the dimension, far exceeding practical needs. We address this gap by analyzing diffusion models through the lens of data covariance spectra, which often follow power-law decays, reflecting the hierarchical structure of real data.
To understand whether such a hierarchical structure can benefit learning in diffusion models, we develop a theoretical framework based on linear neural networks, congruent with a Gaussian hypothesis on the data. We quantify how the hierarchical organization of variance in the data and regularization impacts 
generalization. We find two regimes: When $N <d$, not all directions of variation are present in the training data, which results in a large gap between training and test loss. In this regime, we demonstrate how a strongly hierarchical data structure, as well as regularization and early stopping help to prevent overfitting. For $N > d$, we find that the sampling distributions of linear diffusion models approach their optimum (measured by the Kullback-Leibler divergence) linearly with $d/N$, independent of the specifics of the data distribution.
Our work clarifies how sample complexity governs generalization in a simple model of diffusion-based generative models.

\end{abstract}

\section{Introduction}

\begin{figure*}[!htb]
    \centering
    \includegraphics[width=1.0\linewidth]{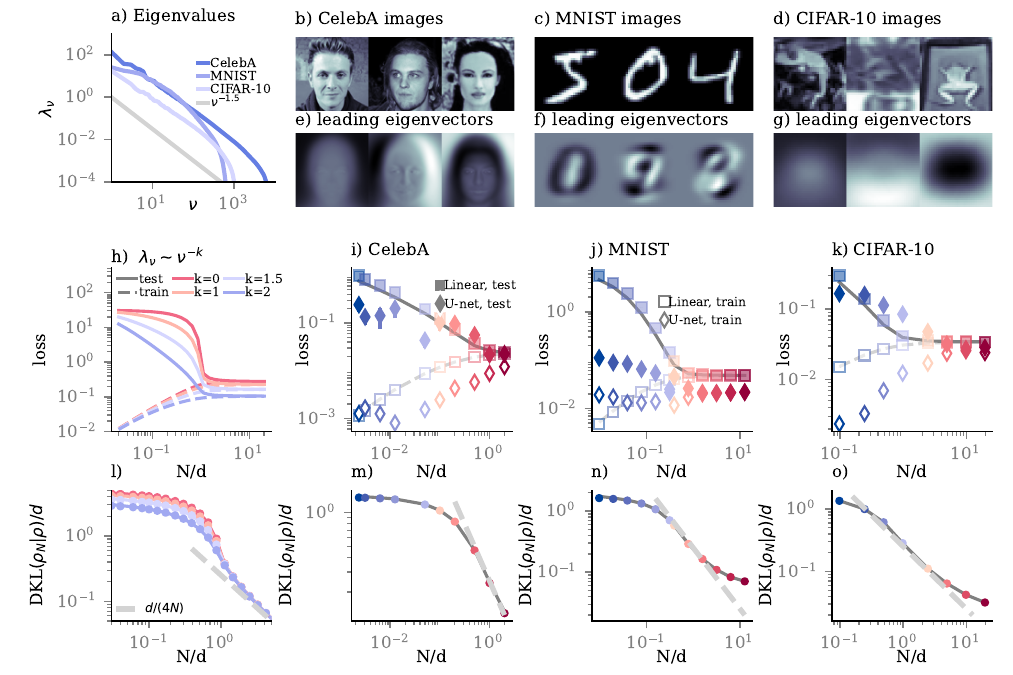}
    \caption{a) Eigenvalues of covariances obtained from image data sets, sorted by rank.  b) - d) Example images from image datasets. e) - g) Top three leading eigenvectors of covariance matrices obtained from the full image dataset. h) Prediction for test and training loss of linear diffusion models trained on $N$ samples $d=100$-dimensional samples from $\mathcal{N}(0,\Sigma)$, where the eigenvalues of $\Sigma$ follow a powerlaw with exponent $k$ normed such that $\Tr \Sigma/d =1$.
    i) - k) Test and train loss of trained diffusion models with linear and U-net architecture trained on $N$ training data. Test losses are averaged over $10^4$ samples from the test set. Training losses are computed using at $\max(N,10^4)$ training data. Grey lines show prediction from replica theory. l) Kullback-Leibler divergence between sample distribution of linear diffusion models with regularization $c=10^{-4}$ and $\mathcal{N}(0,\Sigma)$, where the eigenvalues of $\Sigma$ follow the same powerlaw as in h). Symbols are averages over $10$ random draws of the training sets, error bars report one standard deviation, but are typically smaller than the symbol size. m) - o) are equivalent to l), but for $\Sigma$ originating from the CelebA, MNIST, and CIFAR-10 datasets, respectively. Grey lines show prediction \cref{eq:dkl_from_replica} from replica theory.}
    \label{fig:fig1}
\end{figure*}
Diffusion models \cite{sohl-dickstein_deep_nodate, song_generative_2019, ho_denoising_2020} have become the state-of-the-art paradigm in generative AI, where they are trained to sample from an unknown distribution~$\rho$ based on a
finite set of training data drawn from~$\rho$. 
Ideally, diffusion models abstract the characteristic statistics of $\rho$ from the training data using a neural network. Conventional learning theory dictates that one needs a number of samples $N$ that is exponential in $d$ to accurately approximate a arbitrarily complex function on a $d$-dimensional space, far more than available in practice. The success of diffusion models then appears to indicate that $\rho$ is simple enough such that its key properties can be inferred with fewer data. 

While the behavior of diffusion models that have learned $\rho$ accurately \cite{de_bortoli_diffusion_2021, block_generative_2022,bortoli_convergence_2023, liu_let_2022,lee_convergence_nodate, pidstrigach_score-based_2022} as well as the learning dynamics of diffusion models trained on \textit{unstructured} data (where $\rho$ is the multivariate Gaussian with zero mean and identity covariance) has been studied recently \cite{george_denoising_2025, george_analysis_2025, bonnaire_why_2025}
much less is known about their behavior when trained on finite, \textit{structured} datasets. However, many machine learning datasets exhibit hierarchical structure, where some features are more important than others. Theoretical work on diffusion models usually imposes a particular structure in $\rho$, such as the random hierarchy model \cite{cagnetta_how_2024}, for which generalization occurs when $N$ scales polynomially with $d$, \cite{favero_bigger_2025, favero_how_2025}, though it lacks quantitative predictions for specific datasets. Another line of works assumes data lie on a low-dimensional manifold \cite{chen_score_2023,achilli_memorization_2025,wang_diffusion_2024}, implying sample complexity depends on the manifold dimension rather than the embedding dimension. This manifold hypothesis is ubiquitous in the theory of learning with neural networks \cite{goldt_modeling_2020}. However, estimating manifold dimension is difficult. Moreover, recent work shows a more refined picture of dimensionality in image data \cite{guth_learning_2025} that challenges the notion of these data lying on a lower dimensional manifold of homogeneous dimension in space. 

In contrast to these approaches, we make use of one salient feature of $\rho$ that can be \textit{determined directly from the data}: its covariance spectrum. Covariance spectra, i.e. the distribution of the eigenvalues $\lambda_{\nu}$ of the covariance matrix $\Sigma$ of the data distribution, inform us about the relative spread of the data in different directions. For example, the covariance spectra of image data typically exhibit a power-law behavior, see \cref{fig:fig1} a), meaning that the eigenvalues of $\Sigma$ behave as
\begin{equation}
    \lambda_{\nu} \sim \nu^{-k}, 
    \label{eq:powerlaw}
\end{equation}
where $\nu$ is the rank of the eigenvalue and $k\geq0$ controls level of hierarchy in the data via $k$, where $k=0$ corresponds to a completely flat (non-hierarchical) spectrum, and larger $k$ is associated with a more hierarchical spectrum. This hierarchy in the eigenvalues implies that the few leading eigen-directions of the covariance matrix account for the bulk of the variability in the data. The corresponding eigenvectors are often long-range features of the data, for example controlling background color or the placement of shadows in an image, see \cref{fig:fig1} d)-f), while sub-leading eigenvectors correspond to more detailed features. In the case when $\Sigma$ is exactly known ($N\rightarrow\infty$), covariance spectra are known to affect learning dynamics in diffusion models: directions with larger eigenvalues are typically learned faster than sub-leading ones \cite{catania_theoretical_2025,wang_analytical_nodate}. 

In this work, we investigate how hierarchical covariance spectra affect learning in diffusion models when the number of data, $N$ is finite. In this case, we only have access to $\Sigma$ through a noisy estimate $\Sigma_0$ determined from the data, which grows more accurate with $N$. Moreover, leading eigendirections of $\Sigma$ can typically be recovered at smaller sample complexity, meaning that a more hierarchical structure in $\Sigma$ is beneficial for estimating leading eigendirections from $\Sigma_0$. Here, we investigate how the interplay of the sample complexity $N$, the data hierarchy $k$ and the training dynamics affects generalization dynamics in diffusion models. To this end, we will consider a linear neural network that is able to produce samples with arbitrary covariance. Linear neural networks have helped elucidate overfitting in supervised learning in the past~\cite{advani2020high, fischer_decomposing_2022} by providing a fully tractable case in which key mechanisms can be understood. A first analysis of linear diffusion models was given in \cite{biroli_generative_2023}. Here, we extend their approach to include a hierarchical covariance structure that we fit to data. 
Our main contributions are \vspace{-0.2cm}
\begin{itemize}
    \item When $N$ is smaller than the data dimension $d$, the estimator $\Sigma_0$ has a nullspace of which leads to overfitting. In the same setting, a more hierarchical data model $\Sigma$ benefits generalization.
    \item Regularization mitigates overfitting; the optimal regularization strength decreases with both $N$ and $k$
    \item Larger hierarchy $k$ leads to slower learning and a more gradual increase in test loss, providing a broader window of opportunity for early stopping to mitigate overfitting
    \item changing the standard denoising diffusion objective to predicting the original version of a noised image rather than the noise \textit{reweights} the test loss to be dominated by the leading eigenmodes with increasing $k$ \vspace{-0.2cm}
\end{itemize}
Our theoretical results are complemented by several numerical experiments comparing the behavior of linear models to their nonlinear counterparts. 
We now give a brief introduction to diffusion models before expanding on these results.

\section{Linear Diffusion Models \label{sec:diffusion_models}}
Diffusion models consist of an iterative noising and denoising process. The noising process simplifies the distribution of a sample $x_0$ from $\rho$ through the addition of noise 
\begin{equation}
    x_{t} \left(\epsilon_t \right)= 
    \sqrt{\bar{\alpha}_{t}}x_{0}
    +\sqrt{1-\bar{\alpha}_{t}}\epsilon_{t}
    ,\qquad\epsilon_{t}\sim\mathcal{N}\left(0,\identity\right)\,
    ,\label{eq:x_t_from_x_0}
\end{equation}
for a number of noising steps $t\in\left\{ 1,\dots,T\right\} $ and
$\bar{\alpha}_{t}\in(0,1)$ is a decreasing function in $t$. As $t$ increases, the original signal $x_0$ is gradually suppressed compared to the isotropic Gaussian noise, until one obtains $x_T$ whose distribution is close to $\stdGauss$. Diffusion models are neural networks $\epsilon_{\theta}$ whose parameters $\theta$ are then optimized to approximate the score $\epsilon_{\theta}(x_t,t)\propto \nabla\ln \rho_t(x_t)$ , where $\rho_t$ is the density of the noised variables $x_t$. This is achieved via the denosing score matching objective 
\begin{equation}
\textstyle
    L = \frac{1}{d\,T N} \sum_t \sum_{x_{0} \in \mathcal{D}}
    \mathbb{E}_{\epsilon_t}\left\Vert 
     \epsilon_t 
     -\epsilon_{\theta}
      \left(
       x_{t} \left(\epsilon_t \right),t
      \right)
    \right\Vert^{2}\,,  
    \label{eq:loss}
\end{equation}
where $\mathcal{D}$ is a training set of size $N$ drawn i.i.d. from $\rho$.

Typical architectures for $\epsilon_{\theta}$ are very complex, including U-nets
\cite{ronneberger_u-net_2015} and transformers. Here, we consider the case
where for each $t$, the denoiser $\epsilon_{\theta}(\cdot, t)$ is an affine linear mapping, whose weights we additionally regularize using a standard $L^2$ penalty whose strength $\gamma_t$ depends on $t$.  Once trained, the denoiser is then used to iteratively generate new samples. We provide details on the training and generation process in appendix \ref{app: training_linear_models}.
The learning outcomes of the linear model can be expressed using only the the empirical mean $\mu_0$ and covariance $\Sigma_0$ of the training set $\mathcal{D}$ and the population mean $\mu$ and covariance $\Sigma$ of $\rho$, see \cref{app: training_linear_models}. When the number of training samples, $N$, is finite, the empirical mean and covariance will deviate from their population averages $\mu_0\neq \mu$, $\Sigma_0\neq \Sigma$. The central aim of our work is to predict how this mismatch affects the diffusion model's ability to generalize to new data depending on $k$.

\section{Hierarchical spectra and overfitting}

\paragraph{The nullspace of $\Sigma_0$ drives overfitting.} 
As a first measure for generalization, we now compare the minimal training loss $R$ to the test loss $L_{\text{test}}$, which quantifies the ability of the denoiser to denoise an unseen test example from $\rho$. We express this using the eigendecomposition of the empirical covariance matrix $\Sigma_0$, with eigenvalues $\lambda^0_{\nu}$, and normalized eigenvectors $\{ e^0_{\nu}\}_{\nu}$. 
The gap between the $R$ and $L_{\text{test}}$ is
\begin{equation}
\textstyle
L_{\text{test}}- R 
  = \sum_{t}  \left(\bar{\alpha}_t-\bar{\alpha}^2_{t}\right)
 \sum_{\nu}
  \frac{ \left(e^0_{\nu} \right)^{\T} \Sigma \, e^0_{\nu} -\lambda^0_{\nu}
  +\left(\mu-\mu_{0}\right)_{\nu}^2}{
 \left(
    \bar{\alpha}_{t}\lambda^0_{\nu}
    +\left(1-\bar{\alpha}_{t}+\gamma_{t}\right)
\right)^{2}} 
 \,. \label{eq:test_loss_at_opt}
\end{equation}
This shows shows explicitly that the gap between $R$ and $L_{\text{test}}$ arises whenever there is a mismatch between $\mu_0$, $\Sigma_0$ and $\mu$, $\Sigma$. The terms that contribute the most strongly to $L_{\text{test}}- R$ are those for which the denominator under the sum is minimized. This occurs for the smallest values of $\lambda_{\nu}$ and the smallest values of $t$, as there $1-\bar{\alpha}_t$ is minimized. When the number of data in the training set, $N$, is smaller than the dimension $d$, at least $N-d$ eigenvalues of $\Sigma_0$ are exactly zero, giving rise to large contributions in \cref{eq:test_loss_at_opt}. This is reflected in a very large gap between training and test loss in \cref{fig:fig1} h)-j) when $N<d$. However, this gap can effectively be reduced by regularization, through the presence of $\gamma_t$ in the denominator of \cref{eq:test_loss_at_opt}.
We compare our results using linear models to U-nets trained on image data in \cref{fig:fig1} h) - j). We find a similar saturation at $N\sim d$, but U-nets are naturally able to outperform linear models at large $N$. 

We compute Kullback-Leibler divergence (DKL) comparing the distributions of the generated samples $\rho_N$, and $\rho$. For linear denoisers, $\rho_N = \mathcal{N}\left(\mu_0, \Sigma_0 + c\identity\right)$, where $c$ is a small parameter. The presence of $c$ can be interpreted as originating either from the corrections due to the finite number of sampling steps , or from the regularizing with a particular choice $\gamma_t =\sqrt{\bar{\alpha_t}}c$ for the regularization strength (see \cref{app:sampling_dyn} for detailed derivation). 

We now impose a Gaussian hypothesis on the data $\rho=\mathcal{N}\left(\mu, \Sigma\right)$ to study how the hierarchy in $\Sigma$ controls the deviations between $\rho, \rho_N$ which arise due to finite $N$. The DKL is then
\begin{small}
\begin{align}
\textstyle
    \text{DKL} (\rho_N| \rho) = \frac{1}{2} \Bigg[&
    \ln \frac{\left| \Sigma \right|}{\left| \Sigma_0  + c\identity\right|} 
     + ( \mu - \mu_0 )^{\T} \Sigma^{-1}( \mu - \mu_0 ) \nonumber \\
\textstyle
     &+ \Tr \Sigma^{-1} \left( \Sigma_0  + c\identity\right) - d\Bigg] \,,
     \label{eq:dkl_definition}
\end{align}
\end{small}
which is a measure of distance between distributions. 
The most dominant term at small $N$ is $\text{DKL} (\rho_N| \rho)$ is $\ln \frac{\left| \Sigma \right|}{\left| \Sigma^{\text{eff.}}_0\right|}=\sum_{\nu} \ln \frac{\lambda_{\nu}}{\lambda^0_{\nu} + c}$, where $\lambda_{\nu}$ are the eigenvalues of $\Sigma$. This term in the DKL heavily penalizes the presence of a nullspace in $\Sigma_0$ for small $c$, analogous to the test loss, but also indicates that the penalty for a nullspace of $\Sigma_0$ is less severe when $\Sigma$ has small eigenvalues, hence when its spectrum is more hierarchical.

\paragraph{Hierarchical spectra mitigate overfitting.} So far, our measures of generalization depend on the specific realization of $\Sigma_0$, computed from one draw of the training set. To capture the effect of a hierarchical spectrum of $\Sigma$ explicitly, we now take average over draws of the training set from which $\Sigma_0$ is computed. 
We use the replica trick to evaluate the necessary averages; the full calculation is in found in \cref{app:replica}. For brevity, we will focus on the typical case analysis of the DKL here. Our results depend on a a quantity $q$ which must be determined self-consistently from the eigenvalues $\{ \lambda_{\nu}\}_{\nu=1}^d$ of $\Sigma$ and $c$
\begin{equation}
q =\frac{1}{d}\sum_{\nu}
        \frac{\lambda_{\nu}}{
            1+\lambda_{\nu} \frac{N}{d q +Nc}
            }
        \label{eq: q_main_text}
\end{equation}
 In the average over draws of the training set, the DKL is given by  
\begin{small}
\begin{align}
    \frac{1}{d}\text{DKL} (\rho_N| \rho) =& \frac{1}{2}\frac{q}{\frac{d}{N}q+c}
- \frac{1}{2d}
    \sum_{i}\ln
        \left|
            \frac{c}{\lambda_{i}}
            +\frac{1}{ \frac{d}{Nc}q+1}
        \right| \nonumber\\
&-\frac{N}{2d}\ln\left( \frac{d}{Nc}q+1\right) \nonumber \\
&+\frac{d + 2\sqrt{c} \Tr \Sigma^{-\frac{1}{2}} + c (N+1)\Tr \Sigma^{-1}}{2Nd}
\label{eq:dkl_from_replica}
\end{align}
\end{small}
In \cref{fig:fig1} k) -l) we compare the prediction obtained from \cref{eq:dkl_from_replica} to numerical simulations, showing excellent agreement. 
The first two lines in \cref{eq:dkl_from_replica} originate from the term $\ln |\Sigma| / |\Sigma_0+c\identity|$  and dominate the expression when $c$ is very small. These terms diminish with $q$, hence for smaller $q$, we find that overfitting is less severe at fixed  $N/d$. 
In \cref{app:approximating_q} we also show that 
\begin{equation}
\textstyle
    q \leq \left(\frac{d}{N} \bar{\lambda}
        +c\right)\
        \frac{1}{d} \Tr \frac{\Sigma}{\Sigma+
        \identity \frac{d}{N} \left(\bar{\lambda} +c \right)}
\end{equation}
where $\bar{\lambda}= \frac{1}{d} \Tr \Sigma $ is the average over the eigenvalues of $\Sigma$. At fixed $d/N$, the more hierarchical the spectrum, i.e. the more eigenvalues of $\Sigma$ are significantly smaller than average, the smaller $q$ will be, and therefore, the smaller the DKL. We show examples of this both for spectra which are explicitly powerlaw, $\lambda_{\nu}\sim \nu^{-k}$ and $\Sigma$ determined from image data in \cref{fig:fig1} l)- o). Similar results hold for the gap between test and training loss, shown in the same setting in \ref{fig:fig1} h)-k).  Since \cref{eq:dkl_from_replica} decreases with $q$, this demonstrates how a more hierarchical spectrum can lead to a better fit. 
Intuitively, this is because the absence of variation in $\Sigma_0$ is not as significant when the corresponding variation in $\Sigma$ is also small. 
\begin{wrapfigure}[25]{r}{0.5\linewidth}
    \centering
    \includegraphics[width=\linewidth]{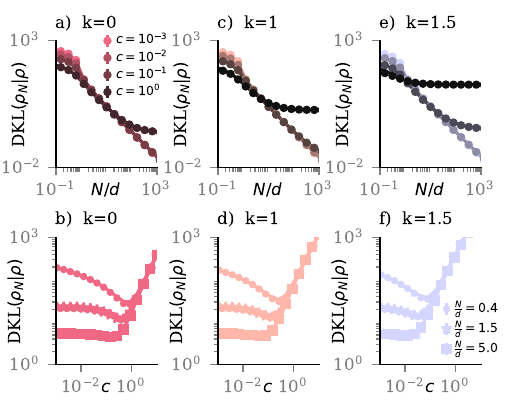}
    \caption{a)-b) Kullback-Leibler divergence between sample distribution of linear diffusion models from Gaussian ground truth  for varying levels of regularization number of data drawn from $\stdGauss$. c)-f) are equivalent to a), b) but for data drawn from and $\mathcal{N}(0,\Sigma)$, where the eigenvalues of $\Sigma$ follow a powerlaw with $\lambda_{\nu}\sim \nu^{-k}$. Symbols are averages over $10$ random draws of the training sets, error bars report one standard deviation, but are typically smaller than the symbol size. Lines show prediction \cref{eq:dkl_from_replica} from replica theory.}
    \label{fig:c_is_cutoff}
\end{wrapfigure}
On the other hand, for $N>d$, the DKL collapses on to the same line independently of the specifics of $\Sigma$. The independence of the DKL on $\Sigma$ has been noted in \cite{tumminello_kullback-leibler_2007}, who argued that this makes the DKL a good measure for the similarity of $\Sigma_0$ to $\Sigma$. In \cref{app:q_in_Nllargerd}, we show that when $c$ is much smaller than the smallest eigenvalue of $\Sigma$ and $N>d$ the DKL is approximately given by $d/(4N)$, where we have neglected terms of order $(d/N)^2$ and $\sqrt{c}$.
In the $N>d$ regime, we find this scaling of the DKL across realizations of $\Sigma$ (see \cref{fig:fig1} k) - l)), up to deviations which originate from $c>0$, which causes a saturation of the DKL above zero.

\paragraph{Regularizing places a spherical bias on the covariance.}


For each timestep $t$, regularizing the linear model with strength $\gamma_t$ effectively replaces $\bar{\alpha}\Sigma_0$ with an effective covariance matrix $\bar{\alpha}\Sigma_0+\gamma_t\identity$. Here, we have chosen to regularize earlier timesteps (which overfit more strongly) more heavily than later timesteps by setting $\gamma_t=\bar{\alpha}_t c$. This results in an effective data covariance matrix $\Sigma_0\rightarrow \Sigma_0+c\identity$. $c$ therefore represents an effective cutoff on the eigenvalues of $\Sigma_0$, because those directions where $\lambda_{\nu}^0 \ll c$ will no longer be distinguishable from each other. 
Hence in the undersampled regime, one can find an optimal $c>0$ as a function of the hierarchy $K$ and $d/N$.

To illlustrate this point, we show the DKL of linear diffusion models trained with different regularization strengths on increasing number of data in \cref{fig:c_is_cutoff} a)-c). For small $N$, regularizing has a beneficial effect. At larger $N$, too much regularization eventually obstructs the structure in $\Sigma$ that could otherwise be resolved via $\Sigma_0$, leading to an increase in the DKL. When the spectrum is more hierarchical, a larger number of eigenvalues is masked by the same level of $c$. This leads to an increase in the DKL of a the regularized linear model from the optimal linear model occurs at smaller $N$ (compare \cref{fig:c_is_cutoff} c) with e)). At fixed, finite number of data, increasing the level of regularization lowers the DKL (improves generalization) up to a certain point, beyond which it is detrimental (see \cref{fig:c_is_cutoff}). The optimal level of regularization decreases when the data becomes more hierarchical, concurrent with our intuition that $c$ masks the subspace $\Sigma_0$ with smaller eigenvalues, which is larger for more hierarchical data.

\paragraph{Regularization and early stopping prevent overfitting.}
\begin{figure*}
    \centering
    \includegraphics[width=\linewidth]{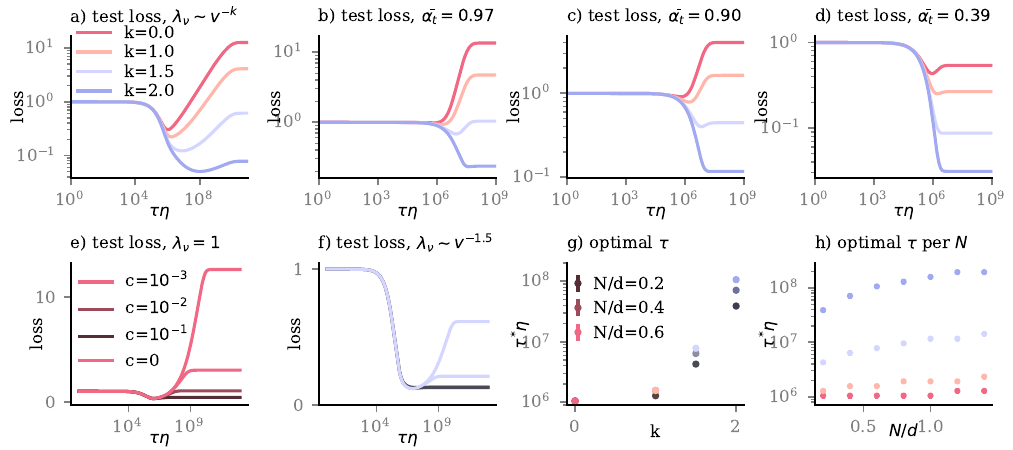}
    \caption{a) Test loss of linear diffusion models trained on $N=0.6d$ samples from a centered Gaussian with powerlaw eigenvalues, $\lambda_{\nu}\sim \nu^{-k}$. b)-d) Are equivalent to b), but for fixed level of noise, $t$. e)-f) are equivalent to b), but for linear denoisers trained with increasing regularization strength $c$. g) Training time $\tau$ with optimal test loss as a function of $k$ for different fractions $N/d$. h) same as g), but for different fractions $N/d$. All curves are averages over $5$ draws of the data and $d=10^3$.}
    \label{fig:Learning_dynamics}
\end{figure*}
We now turn to an analysis of the dynamics of training diffusion models. To simplify the analysis, we assume that the data are centered, $\mu_0=0$ and we start from zero initial weights. Using full batch gradient descent with learning rate $\eta$, one finds that weights of the denoiser exponentially relax to their optima with a direction-and noise dependent rate 
\begin{equation}
\textstyle
    \eta_{\mu,t} =2\frac{\bar{\alpha_{t}}\lambda_{\mu}^{0}+\left(1-\bar{\alpha}_{t}+\gamma_{t}\right)}{dT}\,,
    \label{eq:Mode_time_learning_rate}
\end{equation}
see \cref{app:learning_dyn} for a derivation. This means that leading eigen-directions of $\Sigma_0$ are learned faster than sub-leading ones at fixed $t$, hence the spectral bias observed in diffusion models trained on the population loss \cite{wang_analytical_nodate} translates to diffusion models trained on finite data replacing $\Sigma \rightarrow \Sigma_0$. However, since we know which directions of $\Sigma_0$ are responsible for overfitting, we can now isolate their trajectories to understand how overfitting emerges dynamically during training.
In fact, the gap between training and test loss evolves with the training time for $\tau$ as
\begin{align*}
    \frac{d\left(L(\tau)-R(\tau)\right)}{d\tau}
    \propto
    \sum_{t,\mu} &\left(1-\bar{\alpha}_{t}\right)
    \left(
        1-e^{ -\eta_{\mu,t}\tau} 
    \right) \\
    &\left(\!
        \frac{
        \bar{\alpha}_{t}\left(1- \left(e^0_{\mu} \right)^{\T} \Sigma \, e^0_{\mu}\right)
        +1}{\bar{\alpha_{t}}\lambda_{\mu}^{0}+\left(1-\bar{\alpha}_{t}+\gamma_{t}\right)}-1
        \!\right)\!.
\end{align*}
Starting from initially zero difference between these quantities, modes where 
$\left(e^0_{\mu} \right)^{\T} \Sigma \, e^0_{\mu} > \lambda_{\mu}^{0} +\gamma_t/\bar{\alpha_{t}}$ will make the gap between training and test loss widen over time. Regularization ($\gamma_t>0$ ) can decrease the number of such modes and also shrink the gap. Since $\eta_{\mu,t}\sim \bar{\alpha_{t}}\lambda_{\mu}^{0}+\left(1-\bar{\alpha}_{t}+\gamma_{t}\right)$, the most precarious directions in the sense of overfitting are also the ones which are learned the slowest. This makes both early stopping and regularization effective strategies to prevent overfitting. We show the evolution of the test loss with training time $\tau$ in \cref{fig:Learning_dynamics}a)-d), for specific levels of noise and in the average over all $t$. We find that the optimal $\tau^*$ where the test loss is minimized increases as the spectrum becomes more hierarchical (see \cref{fig:Learning_dynamics}g)) and as we increase $N$ (see \cref{fig:Learning_dynamics}h)). This means both that more hierarchical models require more training, but also overfit later. Regularization prevents overfitting: In \cref{fig:Learning_dynamics} e), f) we apply regularization with strength $\gamma_t=c\bar{\alpha}_t$ to the weights of the model. We see that even moderate levels of regularization can decrease the test loss drastically.

\paragraph{Predicting the data instead re-weights the objective.}
\begin{wrapfigure}[23]{r}{0.5\linewidth}
    \centering
    \includegraphics[width=\linewidth]{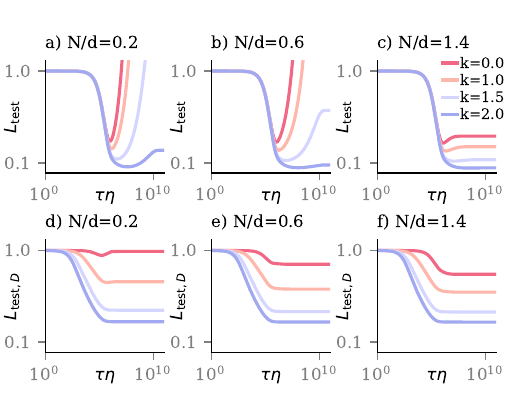}
    \caption{a)-c) Standard test loss of linear diffusion models \cref{eq:loss}, trained on increasing fractions of data as a function of training steps $\tau$ times learning rate $\eta$. d)-e) Same as a)-c), but for test loss corresponding to predicting the data instead of the noise, \cref{eq:loss_on_data}.
    Data are drawn from a centered Gaussian with powerlaw eigenvalues, $\lambda_{\nu}\sim \nu^{-k}$. b)-c)  All curves are averages over $5$ draws of the data and $d=10^3$.}
    \label{fig:Learning_dynamics_ondata}
\end{wrapfigure}
We now investigate whether the learning problem changes if instead
of predicting the noise, one aims to predict the original data. In
principle, both objectives are equivalent, since, if one has a model
$\epsilon_{\theta}(x_{t},t)$ that predicts the noise, one can use
it to define a function $f_{\theta}(x_{t},t)$ to predict the data, 
using that \cref{eq:x_t_from_x_0} implies a linear relation between the two
\begin{align}
f^*_{\theta}(x_{t},t) & =\frac{x_{t}-\sqrt{1-\bar{\alpha}_{t}}\epsilon^*_{\theta}(x_{t},t)}{\sqrt{\bar{\alpha}_{t}}}\,.\label{eq:f_vs_epsilon}
\end{align}
However, we can also choose to directly learn $f_{\theta}$
with the corresponding objective
\begin{equation}
    L_D = \frac{1}{d\,T N} \sum_t \sum_{x_{0} \in \mathcal{D}}
    \mathbb{E}_{\epsilon_t}\left\Vert 
     x_0
     -f_{\theta}
      \left(
       x_{t} \left(\epsilon_t \right),t
      \right)
    \right\Vert^{2}\!,  
    \label{eq:loss_on_data}
\end{equation}
where we use the subscript $D$ to signify we are predicting the data instead of the noise. It is straightforward to show that the optimal affine linear counterpart $f^*_{\theta}$ of $\epsilon^*_{\theta}$ fulfills \cref{{eq:f_vs_epsilon}}. Moreover, 
linear models trained to convergence with either objective have the same Kullback-Leibler divergence and learn modes with the same speed; as we show in \cref{app:data_instead_of_noise}. Consequently, using \cref{eq:loss_on_data} on data rather than \cref{eq:loss} represents a re-weighting of the terms in the loss. Indeed, given a matching initialization, we find that $f_{\theta},  \epsilon_{\theta}$ fulfill \cref{eq:f_vs_epsilon} at any training stage $\tau$. 

This re-weighting of the terms in the test loss is especially significant in the presence of zero modes in $\Sigma_0$. Indeed, we find that the gap between test and training loss for this objective is 
\begin{align}
L_{\text{test},D}- R_D
  =\Tr\left(\Sigma-\Sigma_{0}\right)& \nonumber\\
  +\frac{1}{dT}\sum_{t,\mu}
   &\left[2\left(1-\bar{\alpha}_{t}\right)-\lambda_{\mu}^{0}\bar{\alpha}_{t}\right] \nonumber\\
   & \cdot
\frac{
        \bar{\alpha}_{t}\lambda_{\mu}^{0}
        \left(\left(e_{\mu}^{0}\right)^{\T} \Sigma e_{\mu}^{0}
        - \lambda_{\mu}^{0}\right)}{
        \left(\bar{\alpha}_{t}\lambda_{\nu}^{0}+1-\bar{\alpha}_{t}\right)^{2}}
 \,. \label{eq:test_loss_at_opt_ondata}
\end{align}
where we have neglected terms proportional to the difference of the means $\mu-\mu_0$ for brevity. Here, we see that the terms corresponding to the nullspace of $\Sigma_0$ drop out in the second term, due to the factor $\lambda^0_{\mu}$ in the numerator. Rather, the nullspace of $\Sigma_0$ only contributes to gap between test and training loss via the first term $=\Tr\left(\Sigma-\Sigma_{0}\right)$. Consequently, the objective $L_D$ places comparatively lower emphasis on learning all directions equally well than the objective learning the noise, $L$ (\cref{eq:loss}). 
We show the difference between the two objectives in \cref{fig:Learning_dynamics_ondata} for three different dataset sizes. When $k$ is large, we find that $L_D$ hardly distinguishes between the three cases, whereas we can see a large difference in $L$ and the DKL (see \cref{fig:fig1} l)). For $k=0$, overfitting can be seen by a dynamical decrease and subsequent increase in both objectives when the number of data is small $N/d=0.2$ but is far less pronounced in $L_D$ than in $L$. Thus we see that in the low data regime, $L$ is more representative of the DKL than $L_D$. 

\section{Comparison between linear and non-linear models}

\begin{figure*}[!htb]
    \centering
    \includegraphics[width=\linewidth]{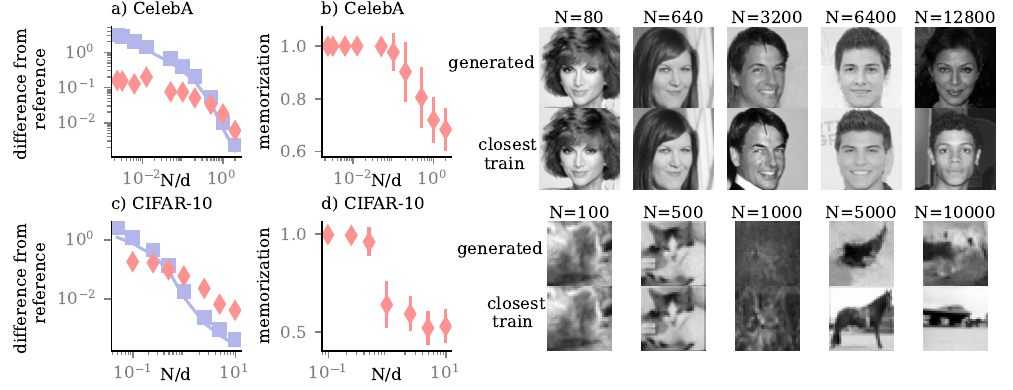}
    \caption{Left column: a) Difference of denoisers from reference model, trained on increasing numbers of data, averaged over $100$ test data points. Blue squares are linear models, pink diamonds are U-nets. Blue lines show prediction from replica calculation. b) Similarity between samples generated from U-net architecture diffusion models and closest training data point, averaged over $400$ generated data points. c)-d) are equivalent to a)-b) but for CIFAR-10 data. Right column: comparison of generated images vs. closest training example for models trained on increasing number of data.\\}
\label{fig:differences_reference}
\end{figure*}
In the previous sections, we focused on understanding how hierarchical spectra influence overfitting and generalization in diffusion models. Now, we shift our focus to understand how and where non-linear models behave differently from linear ones. We will do this in two steps: first, we will investigate the generalization of nonlinear models beyond the test loss. Second, we directly compare the mappings implemented by trained nonlinear models to linear ones.

\paragraph{Convergence to a reference model and memorization.} How close to optimal are models trained on a given dataset size $N$? The test loss in non-linear models shows qualitatively similar behavior to linear models \cref{fig:fig1}), but saturates for both types of models above $N\asymp d$. However, we have seen e.g. from the DKL, that model performance still improves there. For general non-gaussian data and non-linear models, however, the DKL is not accessible. Instead, to measure generalization in a more practical way, we compare a model trained with finite $N$ is to a reference model trained on the population loss. This measure is readily evaluated across datasets and architectures: 
\begin{equation*}
\textstyle
    \Delta \epsilon_N = \frac{1}{T} \sum_t
    \left\langle||\epsilon_{N} (x^{\text{test}}) - \epsilon_{\infty} (x^{\text{test}})_t) ||^2\,\right\rangle_{(x^{\text{test}})},
\end{equation*}
where $(x^{\text{test}})$ are noised test samples, $\epsilon_{N}$ is the mapping obtained from a finite dataset of $N$ samples and $\epsilon_{\infty}$ is the mapping obtained from an infinite number of samples. 
$\Delta \epsilon_N$ compares the mappings implemented by a diffusion model optimized on finite $N$ to the best reference. For linear diffusion models, we find 
\begin{equation}
\textstyle
    \Delta \epsilon_N = 
    L_{\text{test}} -1 + \frac{1}{T} \sum_t 
    \frac{1-\bar{\alpha_{t}}}{(1-\bar{\alpha_{t}}+\gamma_{t})} \Tr\left(\identity+\hat{\alpha}_{t}\Sigma\right)^{-1}\,.
\end{equation}
Note that $\Delta \epsilon_N $ and the test loss \cref{eq:test_loss_at_opt} differ only by terms independent of $N$, which are responsible for the plateau of the test loss. In the unsaturated regime, the test loss is a good proxy for $\Delta \epsilon_N $. For $N>d$, increasing $N$ continues to improve generalization (measured by $\Delta \epsilon_N $), but this effect is no longer reflected in the test loss. In contrast to the test loss, this measure does not saturated above $N\sim d$ and can therefore also measure how non-linear models approach their optima in this regime.

In \cref{fig:differences_reference} a),c) we show $\Delta \epsilon_N$ for two image datasets, both for linear and non-linear diffusion models, as well as a prediction in the average over draws of the dataset from replica theory (for a detailed calculation, see \cref{app:mapping_diff}). In practice, when only finite number of data are available, we choose $\epsilon_{\infty}$ as the mapping optimized on the largest subset of the data. We find that both for linear and non-linear diffusion models, this difference from the reference model decreases significantly at linear sample complexity. 

The most severe consequence of overfitting in diffusion models is \textit{memorization}: when $N$ is small, new samples generated from the trained models can be almost identical to training examples \cite{kadkhodaie_generalization_2024,somepalli_diffusion_2022}. To measure memorization, we find the closest training image for each generated image based on a detail-based similarity measure (defined in \cref{app:SimilarityMeasure}). The similarity between generated and training set examples is highest at small $N$ and decreases as $N$ increases. We find that in our experiments, memorization diminishes when $N\sim d$ and is related to a lower value of $\Delta \epsilon$. However, the samples generated by linear diffusion models are typically more diverse than those of non-linear diffusion models and do not show qualitatively similar behavior. Nevertheless, recent studies highlight early stopping as an effective measure to prevent memorization \cite{favero_bigger_2025,bonnaire_why_2025}, which we observe to be beneficial in linear models as well. 

\paragraph{Similarity between linear and non-linear models.}
\begin{figure*}
    \centering
\includegraphics[width=\textwidth]{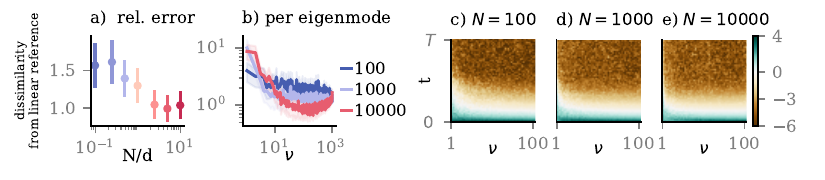}
    \caption{Relative difference of non-linear denoisers from best linear model, per noising step $t$ and direction $\nu$, trained on increasing numbers of data. a) averaged over $\nu$ and $t$, b) averaged over $t$, c) - e) $\log d_{t,\nu}$ per $\nu,t$. All data are averaged over $10^2$ test samples from CIFAR-10 per $t,\nu$.}
\label{fig:differences_linear_nonlinear}
\end{figure*}
Recent experimental studies \cite{wang_hidden_2023, li_understanding_2024} suggest that diffusion models that have learned more general solutions (that are trained on larger number of data) are more similar to linear diffusion models models when $N$ is large. We now test this hypothesis resolved accross levels of noise and directions. We define the following distance measure, evaluated in different eigendirections $\nu$ of $\Sigma$:
\begin{equation}
\textstyle
    d_{t,\nu} = \frac{\left(\epsilon^N\left(x_t,t\right)
    -\epsilon_{\infty}^*\left(x_t,t\right) \right)^2_{\nu}
    }{\left| \left(\epsilon^N\left(x_t,t\right)+\eta 
    \right)_{\nu}
    \left( \epsilon_{\infty}^*\left(x_t,t\right)+\eta 
    \right)_{\nu} \right|}
\end{equation}
where $\epsilon^N$ is a U-net trained on $N$ examples and $\epsilon_{\infty}^*$ is a linear model with modest regularization $c=10^{-2}$, trained on the maximal amount of available data and $\eta=10^{-3}$ prevents divergences. In \cref{fig:differences_linear_nonlinear}, we show $d_{t,\nu}$ for the CIFAR-10 dataset, the same is reported for the CelebA dataset in \cref{app:differences_nonlinear}. Overall, we find that the relative error decays with $N$, $\nu$ and $t$. Indeed for a large extent of $t,\nu$, the difference between linear and non-linear models becomes very small. However, for leading eigenmodes (small $\nu$), the differences between linear and non-linear models grow with $N$. This (small $\nu$, small $t$) is also the regime where we expect the non-Gaussianity of the data to have the largest effect.

\section{Discussion \label{sec:Discussion}}
We have identified two relevant regimes for generalization in linear diffusion
models, $N> d$ and $N< d$. At $N>d$, we observe a saturation in the test loss, and a decay of the Kullback-Leibler divergence that is independent of the data structure. An in-depth treatment of this regime for energy-based models is given by Catania et al. \cite{catania_theoretical_2025}, who show that early stopping and regularization can help mitigate overfitting; the same holds also for linear diffusion models. 
When $N< d$, the model overfits due to a lack of variability in the training set, namely the low-rank structure of the empirical covariance matrix. Both the Kullback-Leibler divergence and test loss strongly penalize this lack of variability. In this regime, a hierarchical data structure that is typical of image data is beneficial for learning. The more hierarchical a dataset is, the lower the test error and Kullback-Leibler divergence will be. 

The presence of regularization in the form of $c$ can be interpreted as placing a cutoff on the minimal variation of the data in any direction, below which the structure of the covariance will no longer be resolved. Hence $c$ plays the role of the relevant scale of the data, a concept that echoes previous investigations relating principal component analysis and deep learning to the renormalization group \cite{mehta_exact_2014, bradde_pca_2017, marchand_multiscale_2023,  kadkhodaie_learning_2023, lempereur_hierarchic_2024}. From a random matrix perspective, it is equivalent to ridge/Tikhonov regression. At finite number of data, one can find an optimal level of regularization that decreases with the level of hierarchy and sample complexity.

Intriguingly, we found that a highly hierarchical structure in the data has no significant effect on the emergence of the two regimes, i.e. the number of data $N$ where these two regimes intersect. This suggests that diffusion models place emphasis on learning all directions with finite variability, not only those with the highest levels of variation. A similar effect has previously been observed in \cite{han_feature_2024}, where learning in the supervised setting was contrasted with diffusion models. In the supervised case, an effect called benign overfitting occurs: if the data consist of a signal that is corrupted by noise, the model may overfit to the signal, ignoring the noise. In diffusion models, however, both the signal and the noise are faithfully represented, meaning that all variability in the data is taken into account and a lack of variability in (any) given direction leads to a large penalty in the test loss. We have shown that this behavior is a consequence of the choice in loss, where one aims to predict the noise instead of the data. This choice has been observed to yield better results in practice \cite{ho_denoising_2020}. On the other hand, predicting the data instead of the noise yields a test loss that emphasizes learning only the leading eigendirections in the data, thus masking the lacking variability in subleading eigendirections. 

Finally, we have shown that the largest difference between linear and non-linear diffusion models occurs in the leading eigendirections at large signal-to-noise ratio, pinpointing the relevant regime for future studies on non-linear diffusion models learning higher order statistics.

%


\section*{Acknowledgements}

We are grateful to Alessio Giorlandino for helpful discussions. CM and SG gratefully acknowledge funding from Next Generation EU, in the context
of the National Recovery and Resilience Plan, Investment PE1 -- Project FAIR
``Future Artificial Intelligence Research’’ (CUP G53C22000440006). SG
additionally acknowledges funding from the European Research Council (ERC) for
the project ``beyond2'', ID 101166056, and from the European
Union--NextGenerationEU, in the framework of the PRIN Project SELF-MADE (code
2022E3WYTY – CUP G53D23000780001).

\clearpage




\newpage
\appendix
\onecolumn
\numberwithin{equation}{section}
\numberwithin{figure}{section}

\section{ Diffusion models \label{app: training_linear_models}}

Diffusion models are designed to reverse the noising process:  they implement a mapping $\epsilon_{\theta}\left( x_t,t\right)$ whose paramaters $\theta$ are optimized to predict the noise vector $\epsilon_t$. For a dataset $\mathcal{D}$ consisting of $N$ samples in $\mathbb{R}^d$, this amounts to minimizing \cref{eq:loss},

which we will refer to as the training loss. 

Here, we consider the case
where for each $t$, the denoiser is implemented by an affine linear mapping 
\begin{equation}
\epsilon_{\theta}\left(x_{t},t\right)=W_{t}\left(x_{t}-\sqrt{\bar{\alpha}_{t}}b_{t}\right)\label{eq:epsilon_affine_linear}
\end{equation}
where $W_t \in \mathbb{R}^{d\times d} $ is a weight matrix and $b_{t} \in \mathbb{R}^{d}$ a bias term. Training the denoiser then amounts to optimizing \cref{eq:loss} with respect to $\left\{W_t,b_t\right\}_{t}$. We will also add a standard regularization term~$\sum_t \gamma_t \Tr W_t W_t^{\T}$ 
to the training objective, where the prefactor $\gamma_t$ allows us to apply different levels of regularization to different noising stages $t$.
With our choice of architecture and regularization, we find the loss to be 
\begin{small}
\begin{equation}
    L = \frac{1}{d\,T |\mathcal{D}|} \sum_t \sum_{x_{0} \in \mathcal{D}}
    \mathbb{E}_{\epsilon_t}\left\Vert 
     \epsilon_t 
     -W_t
      \left(
       \sqrt{\bar{\alpha}_t} x_{t} + \sqrt{1-\bar{\alpha}_t} \epsilon_t
       -\sqrt{\bar{\alpha}_t} b_t
      \right)
    \right\Vert^{2} + \gamma_t \Tr W_t W_t^{\T}\,. 
    \label{eq:loss_linear}
\end{equation}
\end{small}

\subsection{Optimizing the loss \label{app:opt_weights}}

Since $W_t$, $b_t$ are not coupled in \cref{eq:loss} across different values of $t$, we can find their optima independently for each noising stage. Inserting \cref{eq:epsilon_affine_linear} into \cref{eq:loss}, we find that the contribution for each $t$ decomposes into a data-dependent term and one which depends only on the additive noise
\begin{align}
\mathbb{E}_{\epsilon_t}\left\Vert 
     \epsilon_t 
    - \epsilon_{\theta}
      \left(x_t,t
      \right)
    \right\Vert^{2}\, 
   \, \, =\, 
    \bar{\alpha}_{t} 
    \left\Vert 
        W_{t}\left(x_{0}-b_{t}\right)
    \right\Vert^2 
    +\mathbb{E}_{\epsilon_t}\
    \left\Vert 
        \left(1-\sqrt{1-\bar{\alpha}_{t}}W_{t}\right)\epsilon_t
    \right\Vert^2.
  \label{eq:loss_term_averages}
\end{align}
We will not consider variations which arise due to the fact that only a finite number of noised samples are used at each training step, hence in \cref{eq:loss_term_averages} we take the average over infinitely many realizations of $\epsilon_t$.

Eq. \eqref{eq:loss_term_averages} shows that the loss contains only terms either linear or quadratic in the training data $x_0$. Consequently, only the first two moments of the data, or, equivalently, the empirical mean $\mu_0$ and covariance $\Sigma_0$ 
\begin{small}
\begin{align}
    \label{eq:empirical_cumulants}
    &\mu_0 := \frac{1}{N} \sum_{x_{0} \in \mathcal{D}} x_0\,,
    &\Sigma_0 := \frac{1}{N} \sum_{x_{0} \in \mathcal{D}}
                        \left( x_0 - \mu_0 \right)
                        \left( x_0 - \mu_0 \right)^{\T}
            =:\sum_{\nu} \lambda^0_{\nu} e^0_{\nu} \left( e^0_{\nu} \right)^T
\end{align}
\end{small}
of the training data. With this definitions, we now use that \cref{eq:loss_linear} is quadratic in $W_t, b_t$, hence it is convex with the unique minimum 
\begin{small}
\begin{align}
&b_{t}^{*} =\mu_{0}\,,
&W_{t}^{*} =\frac{\sqrt{1-\bar{\alpha}_{t}}}{\bar{\alpha}_{t}\Sigma_{0}+\left(1-\bar{\alpha}_{t}+\gamma_{t}\right)\identity}\,.\label{eq:W_t_star}
\end{align}
\end{small}
At the optimum, $W_t^*, b_t^*$, we find the irreducible, or residual, loss
\begin{small}
\begin{equation}
    R  =\sum_{t}\,\sum_{\nu} \left[\frac{\bar{\alpha}_{t}\lambda^0_{\nu}+\gamma_{t}}{\left(\bar{\alpha}_{t}\lambda^0_{\nu}+\left(1-\bar{\alpha}_{t}+\gamma_{t}\right)\right)}\right]\,.
\label{eq:train_loss_at_opt}
\end{equation}
\end{small}
which is the minimal value of the training loss which a linear denoiser can achieve. 

Once trained, the denoiser is then used to iteratively generate new samples, we provide a description of the generation process in the next section. 

\subsection{The generation process \label{app:generation_process}}
The generation process follows the reverse direction to the noising process, starting from pure white noise $u_0 \sim \stdGauss$ and culminating in a new sample $u_T$. For sampling, we will use $s = T - t$ as an iteration index and $u$ as a dynamical variable to distinguish the noising process from the denoising process. For sampling, we will use $s = T - t$ as an iteration index and $u$ as a dynamical variable to distinguish the noising process from the denoising process. The iteration reads
\begin{equation}
    u_{s+1} =\mu_{\theta}(u_{s},T-s)+\sigma_{T-s} \xi, \qquad\xi \sim\mathcal{N}\left(0,\identity\right)\,.\label{eq:sampling}
\end{equation}
where we defined
\begin{small}    
\begin{equation}
    \mu_{\theta}(x_{t},t) := 
    \frac{x_{t}-\sqrt{1-\bar{\alpha}_{t}\epsilon_{\theta}\left(x_{t},t\right)}}{\sqrt{\alpha_{t}}} 
    + \sqrt{(1-\sigma_{t}^{2})-\bar{\alpha}_{t-1}}   
    \epsilon_{\theta}\left(x_{t},t\right)\,.
    \label{eq:mu_theta}
\end{equation}
\end{small}
These two equations can be understood as first predicting $x_0\approx\frac{x_t-\sqrt{1-\bar{\alpha}_t}\epsilon_{\theta}}{\sqrt{\bar{\alpha}_t}}$, then adding back "noise" in the form of  $\sigma_t\xi+\sqrt{(1-\sigma_{t}^{2})-\bar{\alpha}_{t-1}} \epsilon_{\theta}$.

\section{Sampling dynamics of affine linear denoisers \label{app:sampling_dyn}}
Affine linear denoisers sample from Gaussian distributions: Starting from an initial Gaussian random variable $u_0 \sim \stdGauss$, $u_T$ is a linear map of Gaussian random variables and constants, hence it is also Gaussian. Up to orders of $T^{-1}$, affine linear networks reproduce the mean $\mu_0$ and covariance $\Sigma_0$ of the training data. 
In this section we compute how the mean and covariance of the samples evolve under the the iterative denoising process specified in \cref{eq:sampling}. 
Before we do so, we note that to sample from a given Gaussian distribution with mean $\mu_0$ and covariance $\Sigma_0$, no iterative process is necessary. Rather, given $u_0\sim \stdGauss$, we can generate a sample $u$ with the aforementioned statistics with a simple linear transform 
\begin{equation}
    u_T = \sqrt{\Sigma_0} u_0 + \mu_0 \,.\label{eq:Onestepsampler}
\end{equation}
we will see in the following that the iterative sampling process approaches the same statistics of $u_T$.
In this manuscript, we do not consider the fact that the noising process is discrete, rather we will treat sampling time as continuous. However, we will highlight corrections which arise due to $\bar{\alpha}(0)\neq 1$ and $\bar{\alpha}(T)\neq 0$.

\subsection{Continuous time limit}
\label{app:contin_time}
For the following calculations, it will be useful to write a continuous time version of the denoising process. To this end, we also consider a rescaled time in which the time arguments in the sampling process are incremented by $h=T^{-1},$ rather than increments of $1$ as in \cref{eq:sampling}, thus both denoising time $s$ and noising time $t$ run from zero to one with $t=1-s$. We now assume $\beta(t)=\mathcal{O}(h), \forall t$. This is reasonable, since we expect the changes in every step of the diffusion process to be small. We now define
\begin{align}
&\hat{\beta}(t):=\frac{\beta_t}{h}\, 
&\hat{\sigma}(t) :=\frac{\sigma_t}{\sqrt{h}}
\, .\label{eq:beta_hat}
\end{align}
 For $\bar{\alpha}(t),$we use that $\bar{\alpha}(t+h)=\left(1-\hat{\beta}(t)h\right)\bar{\alpha}(t)$,
so in the limit $T\rightarrow\infty,$ or equivalently $h\rightarrow0$,
\begin{equation}
\frac{d\bar{\alpha}(t)}{dt}=-\hat{\beta}(t)\bar{\alpha}(t)\:.\label{eq:alpha_bar_diffeq}
\end{equation}
 This differential equation admits a formal solution, namely 
\begin{equation}
\bar{\alpha}(t)=e^{-\zeta(t)},\zeta(t):=\int_{0}^{t}ds\,\hat{\beta}(s)\,.\label{eq:gamma_t_def}
\end{equation}
\subsection{Fokker-Planck equation}
With the definition of the continuous noising/denoising time limit in hand, we now write \cref{eq:sampling} as a linear stochastic differential equation. We find \[
du(s)=\left(m(s)u(s)+c(s)\right)ds+\hat{\sigma}(1-s)dZ_{s}
\]
where $m_{s},c_{s}$ are found by inserting \cref{eq:mu_theta} and \cref{eq:W_t_star} into \cref{eq:sampling} and $Z_{s}$ is a Wiener process. We find
\begin{align*}
m(s) & =\frac{\hat{\beta}(1-s)}{2}\identity-\frac{1}{2}\frac{\left(\hat{\sigma}^{2}(1-s)+\hat{\beta}(1-s)\right)}{\bar{\alpha}(1-s)\Sigma_{0}+\left(1-\bar{\alpha}(1-s)\right)\identity},\\
c(s) & =\frac{1}{2}\left(\hat{\sigma}^{2}(1-s)+\hat{\beta}(1-s)\right)\frac{\sqrt{\bar{\alpha}(1-s)}}{\bar{\alpha}(1-s)\Sigma_{0}+\left(1-\bar{\alpha}(1-s)\right)\identity}\mu_{0}\,.
\end{align*}
Observe that $m(s)$ is diagonal in the eigenbasis of $\Sigma_0$. Moving into this basis, we can now solve for the statistics of the sampling process in a decoupled manner, as in this basis, all entries of $u(s)$ are statistically independent. In the following calculation, we will keep one direction $u_{\nu}$ fixed, dropping the index $\nu$ fo brevity.
We  use the Fokker-Planck equation to write down the differential equation for the density $\rho(u,t)$ which describes this variable.
We have 
\begin{equation}
\partial_{s}\rho(u,s)=-\partial_{u}\left[\left(m(s)u(s)+c(s)\right)\rho(u,s)\right]+\frac{1}{2}\partial_{u}^{2}\left[\hat{\sigma}(1-s)^{2}\rho(u,s)\right]\,.\label{eq:Fokker-Planck}
\end{equation}
Since we know that this is a Gaussian process, we make the following Ansatz for the density 
\begin{equation}
\rho(u,s)=\frac{1}{\sqrt{2\pi\sigma_u(s)^2}}
\exp\left[-\frac{\left(u(s)-\mu_{u}(s)\right)^2}{2\sigma_u(s)^2}\right]\nonumber
\end{equation}
defining $\mu_{u}(s), \sigma_u(s)^2$ as the mean and variance of the samples at sampling time $s$, respectively.
With this Ansatz we find that 
\begin{align}
&\partial_{s} (\sigma_u(s)^2)=2m(s)\sigma_u(s)^2+\hat{\sigma}(1-s)^{2}\,
& \partial_{s}\mu_{u}(s)=m(s)\mu_{u}(s)\,+c(s).\label{eq:Diffeq_moments_sample}
\end{align}
which admit the solutions 
\begin{align}
    \mu_u(s) &= \exp\left(\int_0^s dv \, m(v)\right) \left[
    \int_0^s dv \, \exp\left( -\int_0^v dw \, m(w)\right)  c(v)
    + \mu_u(0) \right] \label{eq:sol_sampling_mean} \\
    \sigma_u(s)^2 &= \exp\left(2\int_0^s dv \, m(v)\right) 
    \left[
        \int_0^s dv \, \exp\left( - 2\int_0^v dw \, m(w)\right) \hat{\sigma} (1-v)^2
        + \sigma_u(0)^2 \label{eq:sol_sampling_cov}
    \right]
\end{align}
The initial conditions are $\sigma_u(0)=1$, $\mu_u(0)=0$ since $u(0)\sim \stdGauss$. We will find closed form solutions for these integrals for two choices of $\hat{\sigma}$ in the next step. 

\subsection{Solutions for mean and covariance of samples}
\label{app:sample_solutions}
We will first simplify some of the integrals to solve these two equations.
To find the variance $\sigma_u(s)^2$, we first solve 
\begin{align*}
   \int_0^s dv \hat{\beta}(1-v) & =  \ln \left( \frac{\bar{\alpha}(1-s)}{\bar{\alpha}(1)} \right) \\
   \int_0^s dv \, \frac{-\hat{\beta}(1-v)}{\bar{\alpha}(1-v)(\lambda^0-1)+1}
   &= -\ln \left( \frac{\bar{\alpha}(1-s)}{\bar{\alpha}(1)} \right)
   +\ln \left( \frac{\bar{\alpha}(1-s)(\lambda^0-1)+1}{\bar{\alpha}(1)(\lambda^0-1) + 1} \right) \\
\end{align*}\
Second, we have $c(s)=\sqrt{\bar{\alpha}(1-s)}\left( m(s) - \frac{ \hat{\beta}(1-s)}{2}\right) \mu_0$. With 
\begin{align*}
   & \int_0^s dv \, \exp\left( - \int_0^v dw \, m(w)\right) \sqrt{\bar{\alpha}(1-v)}\left( m(v) - \frac{ \hat{\beta}(1-v)}{2}\right) \\
    =  & \sqrt{\bar{\alpha}(1-s)}\, \exp\left( -\int_0^s dw \, m(w)\right)- \sqrt{\bar{\alpha}(1)}
\end{align*}
we then find 
\begin{align*}
    \mu_u(s) &= \sqrt{\bar{\alpha}(1-s)} \mu_0 - \exp\left(\int_0^s dv \, m(v)\right) \sqrt{\bar{\alpha}(1)}\mu_0 
\end{align*}
We now treat two different scenarios, $\hat{\sigma}^2(t)=\hat{\beta}(t)$ and $\hat{\sigma}(t)=0$.

\subsubsection{$\hat{\sigma}(t)=0$}
In this case, we have $\int_0^s dv \, m(v)=\frac{1}{2} \ln \left( \frac{\bar{\alpha}(1-s)(\lambda^0-1)+1}{\bar{\alpha}(1)(\lambda^0-1) + 1} \right)$, hence 
\begin{align}
    \mu_u(s) &= \sqrt{\bar{\alpha}(1-s)} \mu_0 
    - \sqrt{\bar{
        \alpha}(1)
        \frac{\bar{\alpha}(1-s)(\lambda^0-1)+1}{\bar{\alpha}(1)(\lambda^0-1) + 1}
     }\mu_0 \\
    \sigma_u(s)^2 &= \frac{\bar{\alpha}(1-s)(\lambda^0-1)+1}{\bar{\alpha}(1)(\lambda^0-1) + 1} \nonumber 
\end{align}
Since $\bar{\alpha}(1)$ vanishes exponentially with $\zeta$, and $\bar{\alpha}(0)=1 +\mathcal{O}(h)$, we find that for a long noising trajectory of many steps, $\Sigma_u =\Sigma_0+\mathcal{O}(h)$ and $\mu_u =\mu_0+\mathcal{O}(h)$, reproduce the empirical mean and covariance of the training set to good approximation.

\subsubsection{$\hat{\sigma}^2(t)=\hat{\beta}(t)$}
In this case $\int_0^s dv \, m(v)=\ln \left( \frac{\bar{\alpha}(1-s)(\lambda^0-1)+1}{\bar{\alpha}(1)(\lambda^0-1) + 1} \right)+\frac{1}{2}\ln \left( \frac{\bar{\alpha}(1)}{\bar{\alpha}(1-s)} \right)$, hence we find the sampling mean to be
\begin{align*}
    \mu_u(s) &= \sqrt{\bar{\alpha}(1-s)} \mu_0 
    - \sqrt{\frac{
        \bar{\alpha}(1)^2}{\bar{\alpha}(1-s)}
        \frac{\bar{\alpha}(1-s)(\lambda^0-1)+1}{\bar{\alpha}(1)(\lambda^0-1) + 1}
     }\mu_0 \,.\\
\end{align*}
To evaluate \cref{eq:sol_sampling_cov} for the covariance we first solve the following integral
\begin{align*}
      \int_0^s dv \, \exp\left( - 2\int_0^v dw \, m(w)\right) \hat{\beta} (1-v)
    = &\frac{\left( \bar{\alpha}(1)(\lambda^0-1) + 1\right)^2 }{\bar{\alpha}(1)} \\
    &\cdot
    \left(\frac{\bar{\alpha}(1-s)}{\bar{\alpha}(1-s)(\lambda^0-1)+1} - \frac{\bar{\alpha}(1)}{\bar{\alpha}(1)(\lambda^0-1)+1}  \right) \,.
\end{align*}
Inserting this into \cref{eq:sol_sampling_cov}, we find for the sampling covariance
\begin{align*}
    \sigma_u(s)^2
     =\,\bar{\alpha}(1-s)(\lambda^0-1)+1 +\frac{\bar{\alpha}(1)^2(\lambda^0-1)}{\bar{\alpha}(1-s)} \left( \frac{\bar{\alpha}(1-s)(\lambda^0-1)+1}{\bar{\alpha}(1)(\lambda^0-1) + 1} \right)^2 
\end{align*}\
Again, since $\bar{\alpha}(1)$ vanishes exponentially with $\zeta$, and $\bar{\alpha}(0)=1 +\mathcal{O}(h)$, we find that for a long noising trajectory of many steps, $\Sigma_u(1) =\Sigma_0+\mathcal{O}(h)$ and $\mu_u(1) =\mu_0+\mathcal{O}(h)$, meaning that the sampling mean and covariance reproduce the empirical mean and covariance of the training set to good approximation.

In the case of finite regularization $\gamma_t =c \sqrt{\bar{\alpha}_t}$, we must replace $\lambda^0$ with $\lambda^0+c$ in all formulae. This shows that both regularization $\bar{\alpha}(0)\neq 1$ bias the sampler towards a covariance matrix with an additional, spherical term.

\section{Learning dynamics of linear denoisers}
\label{app:learning_dyn}
Throughout this section, we will assume that all data sets are centered, hence that $\mu_0=\mu=0$ and that the bias terms $b_t$ are initialized at zero, corresponding to their optimal value in this case. 
We introduce a training time $\tau$ and a learning rate $\eta$. At each training step, we will update the parameters of the linear network $\theta$ according to 
\begin{equation*}
    \theta (\tau + d\tau) - \theta (\tau ) d\tau
    = - \eta \nabla_{\theta} L
\end{equation*}
We will treat the dynamics of $W_t$ it in the eigenbasis of the empirical covariance matrix,
\[
W_{t}(\tau)=\sum_{\nu}w_{\mu,\nu,t}(\tau)\,e_{\mu}^{0}\otimes e_{\nu}^{0}\,
\]
Inserting this expression into the training loss, we find 
\begin{equation*}
    L= \frac{1}{dT}\sum_{t}  \sum_{\mu,\nu} 
    \left[
    \left(\bar{\alpha}_{t}\lambda_{\nu}^{0} +\gamma_{t} + 1-\bar{\alpha}_{t}
    \right)
    w_{\mu,\nu,t}^2
    - 2 \sqrt{1-\bar{\alpha}_{t}} w_{\mu,\nu,t} 
    \right]
    +d
\end{equation*}
This expression shows that all entries in $w_{\mu,\nu,t}(\tau)$ decouple and we can treat the evolution of the weight matrices elementwise. Taking the derivative and using the definition of $W_T^*$ (\cref{eq:W_t_star}) we find, that in the limit 
$d\tau \rightarrow 0$
\begin{equation}
w_{\mu,\nu,t}(\tau)=  w_{\mu,\nu,t}^{*}+\exp\left\{ -2\frac{\eta}{dT} \left[\bar{\alpha_{t}}\lambda_{\nu}^{0}+\left(1-\bar{\alpha}_{t}+\gamma_{t}\right)\right]\tau\right\} \left(w_{\mu,\nu,t}(0)-w_{\mu,\nu,t}^{*}\right)\,.
\label{eq:w_dynamics}
\end{equation}
This expression shows that through training, the entries of $W_t$ approach their optimal value exponentially with rate $2\frac{\eta}{dT}\left[\bar{\alpha_{t}}\lambda_{\nu}^{0}+\left(1-\bar{\alpha}_{t}+\gamma_{t}\right)\right]$ corresponding to different spatial directions. The rate $\bar{\alpha_{t}}\lambda_{\nu}^{0}+1-\bar{\alpha}_{t}+\gamma_{t}$ corresponds precisely to the denominator of the terms in \cref{eq:test_loss_at_opt}, whose minimal values lead to the most severe overfitting.

\section{Predicting the data instead of the noise \label{app:data_instead_of_noise}}
In this appendix, we detail the calculations corresponding to the objective \cref{eq:loss_on_data}.  In
principle, both objectives are equivalent, since, if one has a model
$\epsilon_{\theta}(x_{t},t)$ that predicts the noise, one can use
it to define a function $f_{\theta}(x_{t},t)$ to predict the data,
using 
\begin{align}
x_{t} & \hat{=}\sqrt{\bar{\alpha}_{t}}f^*_{\theta}(x_{t},t)+\sqrt{1-\bar{\alpha}_{t}}\epsilon^*_{\theta}(x_{t},t)\label{eq:predicting_x0_vs_noise}\,.
\end{align}
Again, we parametrize the network as an affine linear function 
\[
f_{\theta}(x_{t},t)=V_{t}x_{t}+r_{t}
\]
where $V_{t}$ is a matrix and $r_{t}$ a vector, both are learnable
parameters of the model. Then one finds that the minimum of \cref{eq:loss_on_data} is reached when
\[
r_{t}^{*}=-\left(\sqrt{\bar{\alpha}_{t}}V_{t}^{*}-\identity\right)\mu_{0}\qquad V_{t,\mu\nu}^{*}=\frac{\sqrt{\bar{\alpha}_{t}}\Sigma^{0}}{\bar{\alpha}_{t}\Sigma^{0}+\left(1-\bar{\alpha}_{t}\right)\identity}\,,
\]
which is consistent with \cref{eq:predicting_x0_vs_noise}. Inserting this into the training loss, we find 
\begin{align*}
R & = \frac{1}{dT}\sum_{t}\sum_{\nu}\left(\lambda_{\nu}^{0}-\frac{\bar{\alpha}_{t}\left(\lambda_{\nu}^{0}\right)^{2}}{\bar{\alpha}_{t}\lambda_{\nu}^{0}+1-\bar{\alpha}_{t}}\right) \,,
\end{align*}
which is the irreducible error/ residual loss. The test loss, instead, becomes
\begin{equation}
L_{\text{test},D} =
\frac{1}{dT}\sum_{t}\sum_{\nu}\left(\Sigma_{\nu\nu}-\bar{\alpha}_{t}\left(\lambda_{\nu}^{0}\right)^{2}\left(\frac{\bar{\alpha}_{t}\Sigma_{\nu\nu}+\left(1-\bar{\alpha}_{t}\right)}{\left(\bar{\alpha}_{t}\lambda_{\nu}^{0}+1-\bar{\alpha}_{t}\right)^{2}}\right)-2\bar{\alpha}_{t}\lambda_{\nu}^{0}\frac{\left(1-\bar{\alpha}_{t}\right)\left(\Sigma_{\nu\nu}-\lambda_{\nu}^{0}\right)}{\left(\bar{\alpha}_{t}\lambda_{\nu}^{0}+\left(1-\bar{\alpha}_{t}\right)\right)^{2}}\right)
\end{equation}
For the learning dynamics, assuming
$r_{t}=\mu_{0}=0$ again, one finds 
\begin{align*}
\partial_{\tau}V_{t,\mu\nu}(\tau) & =-\frac{2\eta}{dT}\left(\bar{\alpha}_{t}\lambda_{\nu}^{0}+1-\bar{\alpha}_{t}\right)\left(V_{t,\mu\nu}-V_{t,\mu\nu}^{*}\right)\\
\Rightarrow V_{t,\mu\nu}(\tau) & =V_{t,\mu\nu}^{*}+\exp\left(-\eta_{t,\nu}\tau\right)\left(V_{t,\mu\nu}\left(0\right)-V_{t,\mu\nu}^{*}\right)
\end{align*}
This means that the individual modes of $V_{t}$ are learned at the
same speed as the corresponding modes of $W_{t}$.

\section{Replica theory for linear denoisers at finite $N$ \label{app:replica}}

In this appendix, we derive summary statistics for linear denoisers optimized using the empirical covariance matrix $\Sigma_0$  for different sample sizes $N$. We will assume that the training data originates from a centered Gaussian $\rho \sim \mathcal{N}\left(0,\Sigma\right)$, where we define the "true" covariance matrix $\Sigma$ to be
\begin{equation}
\Sigma=R\Lambda R^{\T},\quad\Lambda=\begin{pmatrix}\lambda_{1} & \dots & 0\\
\vdots & \ddots & \vdots\\
0 & \dots & \lambda_{d}
\end{pmatrix}
\end{equation}
with $R$ a fixed rotation matrix, therefore $|R|=1,R^{\T}R=\identity$.
We parametrize the empirical covariance matrix $\identity+\hat{\alpha}_{t}\Sigma_{0}$
in the following way
\begin{align*}
\Sigma_{0}=\frac{1}{N}\sum_{\beta=1}^{N}\Sigma^{\frac{1}{2}}x^{\beta}\left(\Sigma^{\frac{1}{2}}x^{\beta}\right)^{\T}
,\quad x^{\beta}\sim\stdGauss\:\forall\beta=1,\dots,N
\end{align*}
We are now interested in statistics of the inverse of the related random matrix $\identity + \hat{\alpha}\Sigma_0$. We define the following quantities
\begin{align}
f_g(J)=&
\frac{1}{d}\int\prod_{\beta}d\rho\left(x^{\beta}\right)
\ln Z(J,\Sigma_0)  \nonumber \\
Z(J,\Sigma_0)
:= &\left| \identity + \hat{\alpha} \Sigma_0 +RJ g(\Lambda)R^{\T}\right|^{-\frac{1}{2} } \nonumber  \\
= &\int\frac{d\eta}{\sqrt{2\pi}^{d}}
\exp\left(
    -\frac{1}{2}\eta^{\T}\Bigg[
    \identity+\frac{\hat{\alpha}}{N}\sum_{\beta=1}^{N}\Sigma^{\frac{1}{2}}x^{\beta}\left(\Sigma^{\frac{1}{2}}x^{\beta}\right)^{\T} 
    +RJ g(\Lambda)R^{\T}\Bigg]\eta \right) 
\label{eq:f_definiton}
\end{align}
The function $f$ then plays the role of a generating functional for the moments of
the inverse of $\identity+\hat{\alpha}_{t}\Sigma_{0}$, e.g. 
\[
\frac{1}{d}\left\langle \left(\identity+\hat{\alpha}_{t}\Sigma_{0}\right)^{-1}\right\rangle _{\Sigma_{0}}= R \left(-2\frac{d}{dJ}f(J)|_{J=0,g(\Lambda)=\identity}\right) R^{\T}
\]
For the relevant quantities computed in this manuscript, it will be sufficient to compute $f$ for diagonal $J$, $J_{ij}:= \delta_{ij} J_i$. This is because all quantities can be written as traces of matrix products, and choosing $J$ thus corresponds to choosing the basis in which we evaluate the trace to be given by $R$.

The difficulty in computing $f$ then arises from the fact that all the integrals in $x^{\beta}$ are coupled in the logarithm. To evaluate the integral, we now use the replica trick, which consists of re-writing the logarithm as
\begin{equation}
\left\langle \ln Z(J,\Sigma_0) \right\rangle _{\Sigma_{0}}=\lim_{n\rightarrow0}\frac{1}{n}\left(\left\langle Z(J,\Sigma_0)\right\rangle _{\Sigma_{0}}^{n}-1\right)
\label{eq:replica_trick}
\end{equation}
The replica trick then consists of evaluating $\left\langle Z(J,\Sigma_0)\right\rangle _{\Sigma_{0}}^{n}$ for integer values of $n$ and then taking the limit $n\rightarrow0$.
To compute $\left\langle Z(J,\Sigma_0)\right\rangle _{\Sigma_{0}}^{n}$, we first write the power as an integral over $n$ independent variables
$\eta^{\alpha}$, where $\alpha$ is the replica index, 
\begin{equation}
\left\langle Z (J,\Sigma_0) \right\rangle _{\Sigma_{0}}^{n}=\int\left(\prod_{\alpha=1}^{n}\frac{d\eta^{\alpha}}{\sqrt{2\pi}^{d}}\right)\left\langle \exp\left(\sum_{\alpha=1}^{n}-\frac{1}{2}\left(\eta^{\alpha}\right)^{\T}(\identity+\hat{\alpha}_{t}\Sigma_{0}+RJR^{\T}f(\Sigma))\eta^{\alpha}\right)\right\rangle_{\Sigma_{0}}
\end{equation}
In the following section, we will simplify this expression via a change of variables to a set of summary statistics.

\subsection{Introducing auxiliary variables}

We first isolate the terms depending on $x^{\beta}$ from the expression.  
\begin{align}
\left\langle Z\right\rangle _{\Sigma_{0}}^{n}
=\int
\left(
\prod_{\alpha=1}^{n}\frac{d\eta^{\alpha}}{\sqrt{2\pi}^{d}}
\right) 
& \exp\left(
    \sum_{\alpha=1}^{n}
    \left[
        -\frac{1}{2}\left(\eta^{\alpha}\right)^{\T}\left(\identity+RJf(\Lambda)R^{\T}\right)
    \right]
\right)\nonumber \\
 & \cdot\int
 \left(
 \prod_{\beta=1}^{N}\frac{dx^{\beta}}{\sqrt{2\pi}^{d}}\right)\exp\left(-\frac{1}{2}\sum_{\beta}\left(x^{\beta}\right)^{\T}\left(\frac{\hat{\alpha}}{N}\sum_{\alpha}\Sigma^{\frac{1}{2}}\eta^{\alpha}\left(\Sigma^{\frac{1}{2}}\eta^{\alpha}\right)^{\T}+\identity\right)x^{\beta}
 \right)\label{eq:Z_n}
\end{align}
To simplify the expression a bit, we now change variables to $\mu^{\alpha}=\Sigma^{\frac{1}{2}}\eta^{\alpha}$,
we then find that we can isolate one factor in which $\Sigma$ appears,
but not the samples $x^{\beta}$, and vice versa. Additionally, note that for given
$\mu^{\alpha},$ the second line is just the $N-$th power of the first
line. Both factors are coupled together by the fact that $\mu^{\alpha}$
appear in both factors. We thus simplify to 
\begin{align*}
\left\langle Z\right\rangle _{\Sigma_{0}}^{n}
=\int\left(\prod_{\alpha=1}^{n}
\frac{d\mu^{\alpha}}{\sqrt{2\pi}^{d}}\right) & 
\exp\left(-\frac{1}{2} \sum_{\alpha=1}^{n}\left(R^{\T} \mu^{\alpha}\right)^{\T}
\Lambda^{-\frac{1}{2}}
\left(\identity+J g(\Lambda)\right) 
\Lambda^{-\frac{1}{2}}
\left(R^{\T} \mu^{\alpha}\right)
-\frac{n}{2}\ln|\Lambda|\right)\\
 & \cdot\left[\underbrace{\int\frac{dx}{\sqrt{2\pi}^{d}}\exp\left(-\frac{1}{2}x^{\T}\left(\frac{\hat{\alpha}}{N}\sum_{\alpha}\mu^{\alpha}\left(\mu^{\alpha}\right)^{\T}+\identity\right)x\right)}_{=:G}\right]^{N}
\end{align*}
Another rotation of both $x^{\beta}$ and $\mu^{\alpha}$ by $R^{\T}$ then eliminates $R$ from the expression, leaving all other terms unchanged.
We now simplify the latter integral, $G$. Our goal is to have all
directions $i$ of $\mu^{\alpha}$ decouple. To this end, we define
our first auxiliary variable 
\begin{align}
R_{\alpha} & =\frac{1}{\sqrt{d}}x^{\T}\mu^{\alpha}
\end{align}
and enforce this definition with a Dirac delta in the integral, using that $\delta(r-m)=\frac{1}{2\pi}\int d\tilde{r}\exp\left(i\tilde{r}\left[r-m\right]\right)$.
This yields
\begin{align*}
G= & \int\frac{dx}{\sqrt{2\pi}^{d}}\prod_{\alpha}\frac{dR_{\alpha}d\tilde{R}_{\alpha}}{2\pi}\,\exp\left(-\frac{1}{2}\frac{d\hat{\alpha}}{N}\sum_{\alpha}R_{\alpha}^{2}-\frac{x^{2}}{2}+i\tilde{R_{\alpha}}\left(R_{\alpha}-\frac{1}{\sqrt{d}}x^{\T}\mu^{\alpha}\right)\right)\\
= & \int\prod_{\alpha}dR_{\alpha}\frac{d\tilde{R}_{\alpha}}{2\pi}\exp\left(-\frac{1}{2}\frac{d\hat{\alpha}}{N}\sum_{\alpha}R_{\alpha}^{2}+i\tilde{R_{\alpha}}R_{\alpha}\right)\int\frac{dx}{\sqrt{2\pi}^{d}}\exp\left(-\frac{x^{2}}{2}-i\frac{1}{\sqrt{d}}\tilde{R}_{\alpha}x^{\T}\mu^{\alpha}\right)
\end{align*}
We see that the integral over $x$ is a Gaussian integral which can
be solved exactly, yielding 
\begin{align*}
G & =\int\prod_{\alpha}dR_{\alpha}\frac{d\tilde{R}_{\alpha}}{2\pi}\exp\left(\sum_{\alpha}\left(-\frac{1}{2}\frac{d\hat{\alpha}}{N}R_{\alpha}^{2}+i\tilde{R_{\alpha}}R_{\alpha}\right)-\sum_{\alpha_{1},\alpha_{2}}\frac{\tilde{R}_{\alpha_{1}}\tilde{R}_{\alpha_{2}}}{2d}\left(\mu^{\alpha_{1}}\right)^{\T}\mu^{\alpha_{2}}\right)
\end{align*}
Importantly, this quantity depends only on the replica overlaps $\left(\mu^{\alpha_{1}}\right)^{\T}\mu^{\alpha_{2}}$, which brings us to our second auxiliary variable:
\begin{align}
Q_{\alpha_{1},\alpha_{2}} & :=\frac{1}{d}\left(\mu^{\alpha_{1}}\right)^{\T}\mu^{\alpha_{2}}\label{eq:Q_definition}
\end{align}
Using the Hubbard-Strantonivic transform backwards to also eliminate the integrals over
all $R_{\alpha}$, we find 
\[
G=\int\sqrt{\frac{N}{d\hat{\alpha}}}^{n}\int\prod_{\alpha}\frac{d\tilde{R}_{\alpha}}{\sqrt{2\pi}}\exp\left(-\frac{1}{2}\tilde{R}^{\T}\left(\frac{Q}{d}+\identity\frac{N}{d\hat{\alpha}}\right)R\right)=\sqrt{\frac{N}{d\hat{\alpha}}}^{n}\left|Q+\identity\frac{N}{d\hat{\alpha}}\right|^{-\frac{1}{2}}
\]
Inserting the result for $G$ into the expression as well as enforcing the definition of $Q$ with another Dirac delta, we find 
\begin{align*}
\left\langle Z\right\rangle _{\Sigma_{0}}^{n}
= & \prod_{i}d\rho_{i}(\lambda_{i})\left(\prod_{\alpha_{1}=1}^{n}d\mu^{\alpha_{1}}\prod_{\alpha_{2}=1}^{\alpha_{1}}dQ_{\alpha_{1},\alpha_{2}}\frac{d\tilde{Q}_{\alpha_{1},\alpha_{2}}}{2\pi}\right)\exp\Bigg(S\left(\lambda,\{\mu_{\alpha}\}_{\alpha},Q,\tilde{Q}\right)\Bigg)\\
S\left(\lambda,\{\mu_{\alpha}\}_{\alpha},P,\tilde{P}\right)= & -\frac{1}{2}\sum_{i}\left[\lambda_{i}^{-1}\left(1+J_ig(\lambda_{i})\right)\sum_{\alpha}\left(\mu_{i}^{\alpha}\right)^{2}\right]\\
 & +i\sum_{\alpha_{1}\leq\alpha_{2}}\tilde{Q}_{\alpha_{1},\alpha_{2}}\left(Q_{\alpha_{1},\alpha_{2}}-\frac{1}{d}\left(\mu^{\alpha_{1}}\right)^{\T}\mu^{\alpha_{2}}\right)\\
 & +N\ln G(Q)-\frac{n}{2}\ln|\lambda_{i}|
\end{align*}
Here we have also explicitly used the fact that we chose $J$ to be diagonal in the eigenspace of $\Sigma$.
We now also solve the integral over $\mu^{\alpha}$ by exploiting that all spatial directions in the expression are decoupled. The $\mu^{\alpha}$ dependent part of the integral is then given by
\begin{align}
S_{i}(\tilde{Q}) & =\ln\int 
\prod_{\alpha}\frac{d\mu_{i}^{\alpha}}{\sqrt{2\pi}}
\exp\left(
    -\frac{1}{2\lambda_i}\left(1+J_{i}f(\lambda)\right)\sum_{\alpha}\left(\mu_{i}^{\alpha}\right)^{2}
    -\frac{n}{2}\ln|\lambda_{i}|-\frac{i}{d}
    \sum_{\alpha_{1}\leq\alpha_{2}}
    \tilde{Q}_{\alpha_{1},\alpha_{2}}
    \mu_{i}^{\alpha_{1}}
    \mu_{i}^{\alpha_{2}}
\right)\nonumber\\
 & =\ln \sqrt{\lambda_i}^{-n}\left|\lambda_{i}^{-1}\identity\left(1+J_ig(\lambda_{i})\right)+\frac{i}{d}\,\mathrm{diag}\tilde{Q}+\frac{i}{d}\tilde{Q}\right|^{-\frac{1}{2}}
 \nonumber
\end{align}
With this, we find that the only remaining integrals are in $Q$ and $\tilde Q$. Assuming that $N,\tilde{Q}=\mathcal{O}(d)$, we now pull out a factor $d$
\begin{align}
\bar{Z}^{n}(j)= & \int\prod_{\alpha_{1}\leq\alpha_{2}}^{n}\frac{dQ_{\alpha_{1},\alpha_{2}}d\tilde{Q}_{\alpha_{1},\alpha_{2}}}{2\pi}\exp\left(d S\left(Q,\tilde{Q}\right) \right) \nonumber\\
S\left(Q,\tilde{Q}\right)&=\left[i\sum_{\alpha_{1}\leq\alpha_{2}}\frac{1}{d}\tilde{Q}_{\alpha_{1},\alpha_{2}}Q_{\alpha_{1},\alpha_{2}}+\frac{1}{d}\sum_{i}S_{i}(\tilde{Q})+\frac{N}{d}\ln G(Q)\right]
\end{align}
We will not solve these integrals explicitly. Rather, we will approximate the integral by $\exp\left(d S\left(Q^*,\tilde{Q^*}\right) \right) $, where $Q^*,\tilde{Q^*}$ are the maxima of $S$. This is because due to the common prefactor $d$, the integral is assumed to concentrate around a single point for $d\rightarrow\infty$, the saddle point.

\subsection{Saddle point approximation for any $n$}
Before we find $Q^*,\tilde{Q^*}$, we introduce simplification in the form of a replica symmetric Ansatz 
\begin{align}
Q_{\alpha_{1},\alpha_{2}} & =q\delta_{\alpha_{1},\alpha_{2}}+p(1-\delta_{\alpha_{1},\alpha_{2}})\\
\frac{i}{d}\tilde{Q} & =\tilde{q}\delta_{\alpha_{1},\alpha_{2}}+\tilde{p}(1-\delta_{\alpha_{1},\alpha_{2}})\,.
\end{align}
Parameterizing $Q,\tilde{Q}$ in this way implicitly assumes all replicas are equivalent. With this simplification, we can explicitly diagonalize $Q=npe_{1}e_{1}^{\T}+(q-p)\identity$ with $\,\forall i=1,\dots,n$, where $e_{1}^{\alpha}=\frac{1}{\sqrt{n}}\forall \alpha=1, \dots n$.
Likewise we find $\frac{i}{d}\,\mathrm{diag}\tilde{Q}+\frac{i}{d}\tilde{Q}=n\tilde{p}e_{1}e_{1}^{\T}+(2\tilde{q}-\tilde{p})\identity$.
Inserting this into the matrix determinant, we find
\begin{align*}
G 
 & =\sqrt{\frac{N}{d\hat{\alpha}}}^{n}\left(q-p+\frac{N}{d\hat{\alpha}}\right)^{-\frac{n}{2}}\left(1+\frac{np}{q-p+\frac{N}{d\hat{\alpha}}}\right)^{-\frac{1}{2}}
\end{align*}
and 
\begin{align*}
S_{i}(\tilde{Q}) & =-\frac{n-1}{2}\ln\left|1+J_ig(\lambda_{i})+\lambda_{i}\left(2\tilde{q}-\tilde{p}\right)\right|-\frac{1}{2}\ln\left|1+J_ig(\lambda_{i})+\lambda_{i}\left(2\tilde{q}+(n-1)\tilde{p}\right)\right|-n\ln\lambda_{i}
\end{align*}

We now find and solve the saddle point equations: $\frac{d}{da}S=0$ for $a\in\left\{ q,p,\tilde{q},\tilde{p}\right\} $
for $n\in(0,\infty)$. This yields the following four conditions:
\begin{align*}
\tilde{q} & =\frac{N}{2dn}\left(\frac{(n-1)}{q-p+\frac{N}{d\hat{\alpha}}}+\frac{1}{q+(n-1)p+\frac{N}{d\hat{\alpha}}}\right)\\
\tilde{p} & =\frac{N}{dn}\left(-\frac{1}{q-p+\frac{N}{d\hat{\alpha}}}+\frac{1}{q+(n-1)p+\frac{N}{d\hat{\alpha}}}\right)\\
q & =\frac{1}{dn}\sum_{i}\left[\frac{(n-1)\lambda_{i}}{1+J_ig(\lambda_{i})+\lambda_{i}\left(2\tilde{q}-\tilde{p}\right)}+\frac{\lambda_{i}}{1+J_ig(\lambda_{i})+\lambda_{i}\left(2\tilde{q}+(n-1)\tilde{p}\right)}\right]\\
p & =\frac{1}{nd}\sum_{i}\left[\frac{-\lambda_{i}}{1+J_ig(\lambda_{i})+\lambda_{i}\left(2\tilde{q}-\tilde{p}\right)}+\frac{\lambda_{i}}{1+J_ig(\lambda_{i})+\lambda_{i}\left(2\tilde{q}+(n-1)\tilde{p}\right)}+\right]
\end{align*}
To simplify these expressions, we make the following observations.
First $\tilde{q},\tilde{p}$ only depend on $u:=q-p$ and $w:=q+(n-1)p$.
Second $q,p$ only depend on $\tilde{u}=2\tilde{q}-\tilde{p}$ and
$\tilde{w}=2\tilde{q}+(n-1)\tilde{p}$. It is hence possible to re-parametrize the problem and thereby decouple some of the variables. Concretely, using the first two equations, we find 
\begin{align*}
\tilde{u} & =\frac{N}{d}\frac{1}{u+\frac{N}{d\hat{\alpha}}} &
\tilde{w} & =\frac{N}{d}\left(\frac{1}{w+\frac{N}{d\hat{\alpha}}}\right)
\end{align*}
The second two equations yield 
\begin{align*}
u & =\frac{1}{d}\sum_{i}\left[\frac{\lambda_{i}}{1+J_{i}g(\lambda_{i})+\lambda_{i}\tilde{u}}\right] &
w & =\frac{1}{d}\sum_{i}\left[\frac{\lambda_{i}}{1+J_{i}g(\lambda_{i})+\lambda_{i}\tilde{w}}\right]
\end{align*}
This defines a self-consistency equation each for $u,w$. Interestingly,
both pairs of self-consistency equations are the same. Assuming that the solution
is unique, we find that $u=w$,$\tilde{u}=\tilde{w}$ and hence $p,\tilde{p}=0$.
With this, we find that $q$ must be found self-consistently from
\begin{align}
q & (J)=\frac{1}{d}\sum_{i}\left[\frac{\lambda_{i}}{1+J_{i}g(\lambda_{i})+\lambda_{i}\left(\frac{N}{d}\frac{1}{q+\frac{N}{d\hat{\alpha}}}\right)}\right]\label{eq:Selfconsistq}
\end{align}
and 
\begin{equation*}
    \tilde{q} =\frac{N}{2dn}\frac{1}{q+\frac{N}{d\hat{\alpha}}}
\end{equation*}

\subsection{Taking the limit $n\rightarrow0$}

Inserting the saddle-point values for $q,\tilde{q},p,\tilde{p}$, we find
\begin{align*}
\ln \bar{Z}^{n}(J)= &
dn\Bigg[\frac{N}{2d}\frac{q(J)}{q(J)+\frac{N}{d\hat{\alpha}}}-\frac{1}{2d}\sum_{i}\left[\ln\left|1+J_{i}g(\lambda_{i})+\lambda_{i}\left(\frac{N}{d}\frac{1}{q(J)+\frac{N}{d\hat{\alpha}}}\right)\right|\right] 
\\
& +\frac{N}{2d}\ln\frac{N}{d\hat{\alpha}}-\frac{N}{2d}\ln\left(q(J)+\frac{N}{d\hat{\alpha}}\right)\Bigg]
\end{align*}
Due to the linear appearance of $n$ in $\ln \bar{Z}^{n}(J)$, we can finally take the limit $n\rightarrow0$ in \cref{eq:replica_trick} and find the expression for $f$
\begin{align}
f(J)
= & \frac{N}{2d}\frac{q(J)}{q(J)+\frac{N}{d\hat{\alpha}}}-\left[\frac{1}{2d}\sum_{i}\ln\left|1+J_{i}g(\lambda_{i})+\lambda_{i}\left(\frac{N}{d}\frac{1}{q(J)+\frac{N}{d\hat{\alpha}}}\right)\right|\right]-\frac{N}{2d}\ln\left(\frac{d\hat{\alpha}}{N}q(J)+1\right)\,. 
\label{eq:f_final_J}
\end{align}
Before we move on to take the derivative of $f$ with respect to $J$, we now check that our result is consistent with one result from random matrix theory. To do so, we will first make a relation between $q$ and the following expression
\begin{equation}
   -2\left. \sum_i\frac{df}{dJ_i} \right|_{J=0,g(x)=x} 
   = \frac{1}{d} \left\langle\Tr \Sigma \frac{1}{\identity + \hat{\alpha} \Sigma_0} \right\rangle_{\Sigma_0}
   = \frac{1}{d}\sum_{i}
   \frac{\lambda_{i}}
        {1+\lambda_{i}\left(\frac{N}{d}\frac{1}{q+\frac{N}{d\hat{\alpha}}}\right)} 
    =q
    \label{eq:q_meaning}
\end{equation}
where the first equality follows from the definition of $f$, see \cref{eq:f_definiton}, the second follows from \cref{eq:f_final_J} and the final equality is a consequence of \cref{eq:Selfconsistq}.

\subsection{Consistency checks \label{app:Consistency}}
\begin{figure}[t!]
    \centering
    \includegraphics[width=0.5\linewidth]{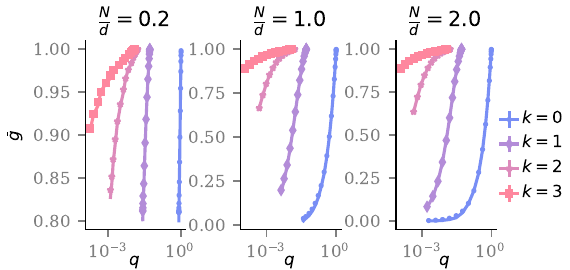}
    \caption{
    Relation between $\bar{g}$ and $q$. Full lines are predictions using \cref{eq:q_barg_relation}, markers show simulations for different fractions of $N/d$, and varying values of $\hat{\alpha}$ between $2^{-9},2^{9}$. Colors and marker styles distinguish and different spectra of the true covariance matrix $\Sigma$,  namely
    $\lambda_i=i^{-k}\,\forall i=1,\dots,d$.
    }
    \label{fig:barg_vs_q}
\end{figure}
To validate our result, we perform a consistency check, using on prior knowledge on the Stieltjes transform $g$ of the random matrix $\Sigma_0=\Sigma^{\frac{1}{2}} \mathcal{W} \Sigma^{\frac{1}{2}} $, defined by
\[
g(z)=\lim_{d\rightarrow\infty}\frac{1}{d}\Tr\frac{1}{z\identity-\Sigma^{\frac{1}{2}}\mathcal{W}\Sigma^{\frac{1}{2}}}
\]
where $\mathcal{W}$ is a Wishart with parameter $\frac{d}{N}$.
It is known that in this case, $g$ fulfills the following self-consistency equation: 
\begin{equation}
g(z)=\int d\rho_{\Sigma}\left(\lambda\right)\frac{1}{z-\lambda\left(1-\frac{d}{N}+\frac{d}{N}zg(z)\right)}\label{eq:Selfconsistg}
\end{equation}
where $\rho_{\Sigma}\left(\lambda\right)$ is the spectral density
of $\Sigma$, $\rho_{\Sigma}\left(\lambda\right)=\frac{1}{d}\sum_{i}\delta(\lambda-\lambda_{i})$
in our case. Defining 
\begin{align*}
\bar{g} & =-\frac{1}{\hat{\alpha}}g\left(-\frac{1}{\hat{\alpha}}\right)=\lim_{d\rightarrow\infty}\frac{1}{d}\Tr\frac{1}{\identity+\hat{\alpha}\Sigma^{\frac{1}{2}}\mathcal{W}\Sigma^{\frac{1}{2}}}
\end{align*}
this variable then fulfills the self-consistency equation 
\begin{equation}
\bar{g}=\frac{1}{d}\sum_{i}\frac{1}{1+\hat{\alpha}\lambda_{i}\left(1-\frac{d}{N}+\frac{d}{N}\bar{g}\right)}\;\label{eq:Selfconsistg_bar} .
\end{equation}
We will now relate the self-consistency relation for $g$ to the self-consistency
equation for $q$. Observe that, from the replica calculation, it
follows that
\[
\bar{g}=\left.\sum_{i}\frac{d}{dJ_i}f(J)\right|_{J=0}=\frac{1}{d}\sum_{i}\frac{1}{1+\lambda_{i}\left(\frac{1}{\hat{\alpha}\frac{d}{N}q+1}\right)}=:s(q)
\]
Using the replica self-consistency relation for $q$, for $J=0,$\cref{eq:Selfconsistq},
we find 
\begin{equation}
s(q):=\frac{\hat{\alpha}\left(\frac{d}{N}-1\right)q+1}{\hat{\alpha}\frac{d}{N}q+1}\,,\label{eq:s_func_definition}
\end{equation}
Furthermore, it follows from the definition \cref{eq:s_func_definition},
that 
\begin{align*}
1-\frac{d}{N}+\frac{d}{N}s(q) & =\frac{1}{\hat{\alpha}\frac{d}{N}q+1}
\end{align*}
which, inserted into the definition of $s$, yields a self-consistency
equation for $s(q)$, 
\begin{equation}
\label{eq:sq_selfconsist}
s(q)=\frac{1}{d}\sum_{i}\frac{1}{1+\hat{\alpha}\lambda_{i}\left(1-\frac{d}{N}+\frac{d}{N}s(q)\right)}
\end{equation}
which is identical to the self-consistency equation for $\bar{g}$,
\cref{eq:Selfconsistg_bar}, meaning that the replica calculation
produces a variant of the known self-consistency relation \cref{eq:Selfconsistg} for the Stieltjes transform. Furthermore the replica calculation yields a relation between $q=\frac{1}{d}\left\langle \Tr\left(\Sigma\frac{1}{1+\hat{\alpha}\Sigma_{0}}\right)\right\rangle _{\Sigma_{0}}$and
$\bar{g}=\frac{1}{d}\left\langle \Tr\left(\frac{1}{1+\hat{\alpha}\Sigma_{0}}\right)\right\rangle _{\Sigma_{0}}$ which
is independent of the spectrum of $\Sigma$, namely 
\begin{equation}
   \bar{g} = s(q)\label{eq:q_barg_relation}
\end{equation}
We test this relation in \cref{fig:barg_vs_q}, finding excellent agreement with simulations. 

\subsection{ Squared difference on test examples \label{app:mapping_diff}}

We compute the generalization measure for fixed $t$:
\begin{align*}
\langle\left|\left|\left(W_{t}^{*}-W_{t}^{\text{oracle}}\right)x_{t}\right|\right|^{2}\rangle_{x_{t},\Sigma_{0}} & =\frac{1-\bar{\alpha_{t}}}{(1-\bar{\alpha_{t}}+\gamma_{t})}\left(\frac{\psi_{1,1} + \psi_{1,2}}{(1-\bar{\alpha_{t}}+\gamma_{t})} -2 \psi_2 + \psi_3 \right)
\end{align*}
where we defined
\begin{align}
&\psi_{1,1}^t  = \left\langle \Tr\left[\frac{1-\bar{\alpha_{t}}}{\left(\identity+\hat{\alpha}_{t}\Sigma_{0}\right)^{2}}\right] \right\rangle
&\psi_{1,2}^t  = \left\langle  \Tr\left[\frac{\bar{\alpha_{t}}}{\left(\identity+\hat{\alpha}_{t}\Sigma_{0}\right)^{2}}\Sigma\right] \right\rangle\quad
&\psi_{2}^t =\left\langle  \Tr\left[\frac{1}{\left(\identity+\hat{\alpha}_{t}\Sigma_{0}\right)}\right]\right\rangle
 \label{eq:psi_definitions}
\end{align}
We first compute $\psi_{2}$, using that
\[
\psi_{2}=-2 \sum_{i}\frac{d}{dJ_i}f(J)\Bigg|_{J=0,g=1}
\]
We now find $\frac{\partial f(J)}{\partial q}=0$, hence $\frac{df}{dJ_i}=\frac{\partial f(J)}{\partial J_i}$,
which is equal to 
\[
\frac{df}{dJ_i}=-\frac{1}{2d}\frac{g(\lambda_{i})}{1+J_i g(\lambda_{i})+\lambda_{i}\left(\frac{N}{d}\frac{1}{q+\frac{N}{d\hat{\alpha}}}\right)}
\]
Where $q$ is found, e.g. by solving \cref{eq:Selfconsistg_bar}. 

For $\psi_{1,1}$and $\psi_{1,2}$, we first note that for precision
matrix $A$, and diagonal $J,\Lambda$, we have that 
\begin{align*}
\frac{d^2}{dJ_idJ_j} \left.\ln\int\frac{d\eta}{\sqrt{2\pi}^{d}}\exp\left(-\frac{1}{2}\eta^{\T}\left[A+J g(\Lambda)\right]\eta\right)\right|_{J=0} 
 & =\frac{f(\Lambda)_{ii}f(\Lambda)_{jj}}{2}\left(A_{ij}^{-1}\right)^{2}
\end{align*}
Where we used Wick's theorem to evaluate the Gaussian moments. For
the specific functions of interest \cref{eq:psi_definitions},
we find
\begin{align*}
&\psi_{1,1}=  2
\sum_{ij}\left(\frac{d^{2}}{dJ_idJ_j}f(J)\right)\Bigg|_{J=0,g(x)=1}
&\psi_{1,2}=  2
\sum_{ij}\left(\frac{d^{2}}{dJ_idJ_j}f(J)\right)\Bigg|_{J=0,g(x)=\sqrt{x}}
\end{align*}
the only difference between the two being the function $g$. Hence, we compute the second derivatives
\[
\left.\frac{d^{2}f}{dJ_idJ_j}\right|_{J=0}
=\left.\delta_{ij}\
\frac{1}{2d}\frac{g^{2}(\lambda_{i})}{\left(1+\lambda_{i}\left(\frac{N}{d}\frac{1}{q+\frac{N}{d\hat{\alpha}}}\right)\right)^{2}}
-\frac{N}{2d^{2}}\frac{g(\lambda_{i})\lambda_{i}}{\left(q+\frac{N}{d\hat{\alpha}}+\frac{N}{d}\lambda_{i}\right)^{2}}
\frac{dq}{dJ_j}\right|_{J=0}
\]
Using the self-consistency equation, we find 
\begin{align*}
\left.\frac{dq}{dJ_i}\right|_{J=0} & 
=-\frac{\left(q+\frac{N}{d\hat{\alpha}}\right)^{2}}
{d\left(1-\frac{N}{d}R_{2}(q)\right)}
\frac{\lambda_{i}g(\lambda_{i})}
{\left(q+\frac{N}{d\hat{\alpha}}+\frac{N}{d}\lambda_{i}\right)^{2}}\\
\end{align*}
where we defined 
\begin{equation*}
    R_{k}(q)  =\frac{1}{d}\sum_{j}\frac{\lambda_{j}^{k}}{\left(q+\frac{N}{d\hat{\alpha}}+\frac{N}{d}\lambda_{j}\right)^{2}}\,.
\end{equation*}
Putting it all together, for the specific functions of interest, we find 
\begin{align*}
&\psi_{1,1}=  \left(q+\frac{N}{d\hat{\alpha}}\right)^{2}\left[R_{0}+\frac{N}{d}\frac{R_{1}(q)^{2}}{\left(1-\frac{N}{d}R_{2}(q)\right)}\right] 
&\psi_{1,2}  =\left(q+\frac{N}{d\hat{\alpha}}\right)^{2}\left[R_{1}+\frac{N}{d}\frac{R_{\frac{3}{2}}(q)^{2}}{\left(1-\frac{N}{d}R_{2}(q)\right)}\right] \nonumber \\
\end{align*}
and 
\begin{equation*}
    \psi_{2} =\frac{1}{d}\sum_{i}\frac{1}{1+\lambda_{i}\left(\frac{N}{d}\frac{1}{q+\frac{N}{d\hat{\alpha}}}\right)} \,.
\end{equation*}

\subsection{Residual and test loss at finite $N$ \label{app:test_train_loss}}

To compute the residual loss, we employ the following identity:
\begin{align*}
\left.\frac{df}{d\hat{\alpha}_{t}}\right|_{J=0,g=1}
&=-\frac{1}{2Nd}\sum_{\beta=1}^{N}\left\langle \sum_{ij}\left(\Sigma^{\frac{1}{2}}x^{\beta}\right)_{i}\left(\frac{1}{1+\hat{\alpha}\Sigma_{0}}\right)_{ij}\left(\Sigma^{\frac{1}{2}}x^{\beta}\right)_{j}\right\rangle \\
&= -\frac{1}{2d}\left\langle\Tr \Sigma_0 \frac{1}{\identity+\hat{\alpha}\Sigma_0}\right\rangle 
\end{align*}
Inserting this into the equation for the residual loss, \cref{eq:train_loss_at_opt}, which yields 
\begin{align}
    R =\frac{-2}{T} \sum_{t} &\,\left. 
    \frac{\bar{\alpha}_{t}}{1-\bar{\alpha_{t}}+\gamma_{t}}\frac{df}{d\hat{\alpha}_{t}} \right|_{J=0,g=1}
    +
    \frac{\gamma_{t}}{1-\bar{\alpha_{t}}+\gamma_{t}} \psi_2 \nonumber \\
    = \frac{1}{T} \sum_{t}\, &
    \frac{\bar{\alpha}_{t}}{1-\bar{\alpha_{t}}+\gamma_{t}} \frac{q}{\frac{d\hat{\alpha}_t}{N}q+1}
    +\frac{\gamma_{t}}{1-\bar{\alpha_{t}}+\gamma_{t}}
    \frac{1}{d} \sum_i \frac{1}{1+\lambda_{i}\left(\frac{N}{d}\frac{1}{q+\frac{N}{d\hat{\alpha}_t}}\right)}
\end{align}
Second, we find that the test loss simplifies to
\[
L_{test}=1+\frac{1}{T}\sum_{t}
\frac{1-\bar{\alpha_{t}}}{(1-\bar{\alpha_{t}}+\gamma_{t})}\left(\frac{\psi_{1,1} + \psi_{1,2}}{(1-\bar{\alpha_{t}}+\gamma_{t})} -2 \psi_2 \right) 
\]

which contains only functions which we have already computed in \cref{app:mapping_diff}.

\subsection{Kullback-Leibler divergence \label{app:Kullback-leibler}}

We compare $\rho = \mathcal{N} \left( \mu, \Sigma \right)$ to $\rho_N = \mathcal{N} \left( \mu_0, \Sigma_0 +\gamma \identity \right)$
The DKL between two Gaussians is given by 
\begin{equation}
    \text{DKL} (\rho_N| \rho) = \frac{1}{2} \left[
    \ln \frac{\left| \Sigma \right|}{\left| \Sigma_0  + c\identity\right|}
     + ( \mu - \mu_0 )^{\T} \Sigma^{-1}( \mu - \mu_0 ) + \Tr \Sigma^{-1} \left( \Sigma_0  + c\identity\right) - d\right] \,,
\end{equation}
We now average this expression over draws of the data set term by term. First, note that 
\begin{equation}
    \Tr \Sigma^{-1} \left( \langle \Sigma_0 \rangle + c \identity \right) - d
    = c \, \Tr \Sigma^{-1}
\end{equation}
Second, we compute 
\begin{align*}
   \left\langle  ( \mu - \mu_0 )^{\T} \Sigma^{-1}( \mu - \mu_0 ) \right\rangle
    &= \frac{1}{N^2}\sum_{i,j,k,l=1}^d\sum_{\beta_1,\beta_2=1}^N 
     \left\langle x_i^{\beta_1} x_j^{\beta_2} \right\rangle \left(\Sigma^{\frac{1}{2}} + \sqrt{c} \identity\right)_{ki} \Sigma^{-1}_{kl} \left(\Sigma^{\frac{1}{2}} +\sqrt{c} \identity \right)_{lj} \\
    &=\frac{d + 2\sqrt{c} \Tr \Sigma^{-\frac{1}{2}} + c \Tr \Sigma^{-1} }{N}
\end{align*}
with $x^{\beta}\sim\stdGauss$. Finally, we have that when $c= \frac{1}{\hat{\alpha}}$
\begin{equation}
    -\frac{1}{2}\left\langle \ln \left| \Sigma_0 +c \identity \right|\right\rangle 
    = d \, f(J=0) + \frac{d}{2} \ln \hat{\alpha}
\end{equation}

we now evaluate \cref{eq:f_final_J} at $J=0$ to find 

\begin{align*}
    -\frac{1}{2}\left\langle \ln \left| \Sigma_0 + \hat{\alpha}^{-1} \identity \right|\right\rangle 
= &  \frac{N}{2d}\frac{q}{q+\frac{N}{d\hat{\alpha}}}
-\left[\frac{1}{2d} 
\sum_{i}\ln\left|\frac{1}{\hat{\alpha}}+\lambda_{i}\left(\frac{N}{d}\frac{1}{q\hat{\alpha}+\frac{N}{d}}\right)\right|\right]-\frac{N}{2d}\ln\left(\frac{d\hat{\alpha}}{N}q+1\right) \\
\end{align*}
such that all in all, the DKL simplifies to \cref{eq:dkl_from_replica}. 

\subsection{Bounds and approximations of $q$ \label{app:approximating_q}}

\subsubsection{Bounding $q$ from above}

\begin{figure}
    \centering
    \includegraphics[width=\linewidth]{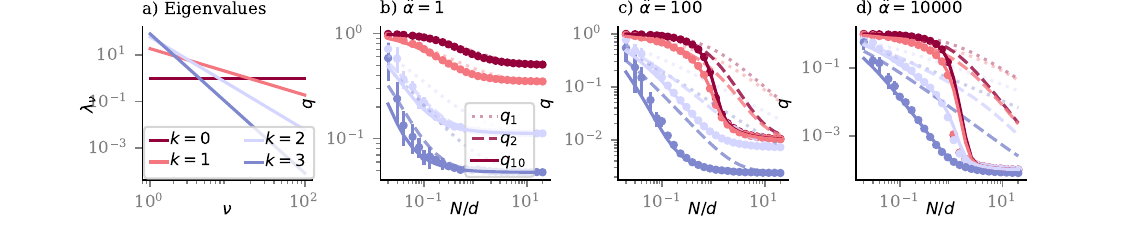}
    \caption{Comparison of $q$ to the upper bound. a) shows the spectra of the covariances we consider, $\lambda_{\nu}=c_k \nu^{-k} $ where we choose $c_k$ such that $\frac{1}{d}\sum_{\nu}\lambda_{\nu}=1$. b)-f) compare numerical values of $q$ found using \cref{eq:q_meaning} with $d=100$. Dots and error bars denote mean and standard deviations over ten realizations of $\Sigma_0$, respectively. Dotted, dashed and full lines show upper bounds \cref{eq:q_bound_n} for increasing $n$.}
    \label{fig:q_bound}
\end{figure}

We now use two different approaches to bound $q$ from above. Defining 
\begin{equation*}
    h(q) = \frac{1}{\frac{d}{N}q+\frac{1}{\hat{\alpha}}} \,
\end{equation*}
we find that 
\begin{equation*}
    q=\frac{1}{d}\sum_{\nu=1}^{d}\frac{\lambda_{\nu}}{h(q)\lambda_{v}+1}
    = h^{-1}(q) \frac{1}{d}\sum_{\nu=1}^{d}\frac{\lambda_{\nu}}{\lambda_{v}+h^{-1}(q)}
\end{equation*}
First, note that from \cref{eq:q_meaning} follows that $q > 0$ hence, $h^{-1}(q) >0$. With this, we can make a very coarse approximation that 
\begin{equation*}
    q\geq\frac{1}{h(q)} \Rightarrow\left(1-\frac{d}{N}\right)q\geq\frac{1}{\hat{\alpha}}
\end{equation*}
For $d<N$, this bounds $q$ from above via 
\begin{equation}
    d<N \quad \Rightarrow \quad q \leq\frac{1}{\hat{\alpha}\left(1-\frac{d}{N}\right)} \label{eq: q_bound_1}
\end{equation}
For $\hat{\alpha}$ very large, one can show that the difference to the right hand side (see \cref{app:q_in_Nllargerd}) is of order $\mathcal{O}\left(\hat{\alpha}^{-2}\right)$. 
This bound is only valid when $N>d$, additionally, it does not depend on the dimension. When $N<d$, we may instead use 
\begin{equation}
    q\leq  \frac{1}{d} \Tr \Sigma
\end{equation}
this approximation itself is quite coarse. However, we may reinsert it into the equation for $q$ to obtain a smaller upper bound
\begin{equation*}
    h(q)\geq\frac{1}{\frac{d}{N}\frac{1}{d} \Tr \Sigma +\frac{1}{\hat{\alpha}}} 
    \Rightarrow 
    q \leq \left(\frac{d}{N} \frac{1}{d} \Tr \Sigma 
    +\frac{1}{\hat{\alpha}}\right)\
    \frac{1}{d} \Tr \frac{\Sigma}{\Sigma+
    \identity  \left(\frac{d}{N}\frac{1}{d} \Tr \Sigma +\frac{1}{\hat{\alpha}} \right)}
\end{equation*}
which is a slightly tighter bound on $q$, which we can reinsert into the expression for $q$ again to obtain an even smaller upper bound. We may alternatively define the following series 
\begin{equation}
    q_0 = \frac{1}{d} \Tr \Sigma, \quad q_{n+1} = 
    \left(\frac{d}{N} q_n 
    +\frac{1}{\hat{\alpha}}\right)\
    \frac{1}{d} \Tr \frac{\Sigma}{\Sigma+
    \identity \left( \frac{d}{N} q_n +\frac{1}{\hat{\alpha}} \right)}
    \label{eq:q_bound_n}
\end{equation}
we then find that $q\leq q_n \forall n$. We compare numerical simulations of \cref{eq:q_meaning} to the bound on $q$ obtained thus in \cref{fig:q_bound}, finding that the bound becomes increasingly tight as we increase $n$. 

\subsubsection{$q$ at $N\gg d$ and large $\hat{\alpha}$ \label{app:q_in_Nllargerd}}

We now seek an approximation for $q$ in the regime $N\gg d$ and large $\hat{\alpha}$. To do so, we first examine $\bar{g}=\frac{1}{d}\left\langle \Tr\left(\frac{1}{1+\hat{\alpha}\Sigma_{0}}\right)\right\rangle _{\Sigma_{0}}$, which we found to relate to $q$ via the relation \cref{eq:q_barg_relation}. We now additionally assume that $\hat{\alpha}$ is much larger than $\left(\lambda^0_{\text{min}}\right)^{-1}$, where $\lambda^0_{\text{min}}$ is the smallest eigenvalue in $\Sigma_0$. This assumption is only valid for at least $N\geq d$ as otherwise $\Sigma_0$ has zero eigenvalues.
Then it follows that $\bar{g}$ is of order $\hat{\alpha}^{-1}$. We now invert the relation between $q$ and $\bar{g}$, finding that 
\begin{equation}
    q =\frac{1}{\hat{\alpha}\frac{d}{N}}\left(\frac{1}{1-\frac{d}{N} +\frac{d}{N} \bar{g}}-1\right) \approx \frac{1}{\hat{\alpha}\frac{d}{N}}\left(\frac{1}{1-\frac{d}{N}}-1\right)
\end{equation}
which is independent of $\Sigma$. Inserting this into \cref{eq:dkl_from_replica}, we find 
\begin{equation}
    \frac{\text{DKL} (\rho_N| \rho)}{d}
    = \frac{1+\frac{N}{d}\ln(1-\frac{d}{N}) +\ln(1-\frac{d}{N})+\frac{N}{d^2}}{2} + \mathcal{O} \left( \sqrt{\hat{\alpha}^{-1}}\right)\,,
\end{equation}
which, for $N \gg d $, scales as $\frac{d}{4N}$.

\section{A detail-based similarity measure for CelebA and CIFAR-10 \label{app:SimilarityMeasure}}

For the models trained on image data, computing the cosine similarity $c(x,y)=\frac{x^{\T}y}{|x||y|}$ between a generated sample and one from the training set yields a very high similarity $\sim 0.9$, even if the generated images are genuinely different. Upon manual inspection of the corresponding images, we find that this occurs due to a large portion of the image, such as the background, being uniformly dark or light. 

\begin{figure}
    \centering
    \includegraphics[width=\linewidth]{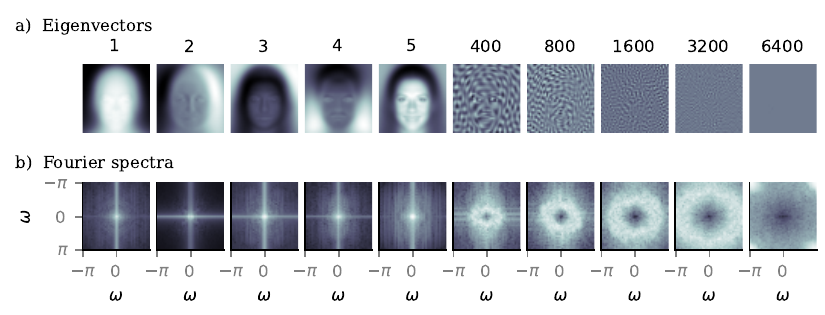}
    \caption{a) First five leading eigenvectors $\nu =1,\dots, 5$ of the covariance matrix of the CelebA data, as well as sub-leading eigenvectors $\nu = 400,\dots, 6400$. b) Corresponding Fourier spectra of the eigenvectors.}
    \label{fig:eigenvectors-fourier}
\end{figure}

In \cref{fig:eigenvectors-fourier} we compare the eigenvectors of the covariance matrix of the CelebA data to their corresponding Fourier spectra. We find that leading eigenvectors $\nu=1,\dots, 5$ have a more homogeneous spatial distribution of light and dark pixels, correspondingly their Fourier spectra are concentrated around small frequencies (small $|\omega|$). As $\nu$ increases, however, the spectra of the eigenvectors become more broad, and small frequencies are suppressed. 

On the basis of these observations, we construct a similarity measure which is oriented more towards the details of the images: 
We first project the images into the space spanned only by sub-leading eigenvectors $\nu >5$. We then compute the cosine similarities of the resulting vectors. We find that this measure is then more sensitive to changes in the details of the images, which leave the background uniform (e.g. for the generated image and closest training set examples in \cref{fig:fig1} for $N\geq 6400$).

\section{Differences between linear and non-linear models \label{app:differences_nonlinear}}
\begin{figure}
    \centering
    \includegraphics[width=\linewidth]{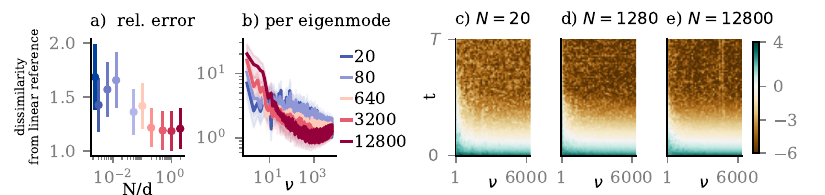}
    \caption{Relative difference of non-linear denoisers from best linear model, per noising step $t$ and direction $\nu$, trained on inecreasing numbers of data. a) averaged over $\nu$ and $t$, b) averaged over $t$, c) - e) $\log d_{t,\nu}$ per $\nu,t$. All data are averaged over $100$ test samples per $t,\nu$.}
    \label{fig:differences_linear_nonlinear_celeba}
\end{figure}
In \cref{fig:differences_linear_nonlinear_celeba} we report the difference in the mapping of the CelebA compared to the best linear model. We observe qualitatively the same behavior as in the CIFAR-10 data: overall, the relative error decays with $N$, $\nu$ and $t$. Indeed for a large extent of $t,\nu$, the difference between linear and non-linear models becomes very small. However, for leading eigenmodes (small $\nu$), the differences between linear and non-linear models grow with $N$.

\end{document}